\documentclass[runningheads]{llncs}
\pdfoutput=1
 
\usepackage{eccv}

\usepackage{eccvabbrv}

\usepackage{graphicx}
\usepackage{booktabs}
\usepackage[dvipsnames]{xcolor}
\usepackage{amsfonts}
\usepackage{amsmath}
\usepackage{multirow}
\usepackage{indentfirst}
\usepackage{array}
\usepackage{xltabular}
\usepackage{wrapfig}
\usepackage{soul}

\DeclareMathOperator*{\argmin}{argmin}

\newcolumntype{H}{>{\setbox0=\hbox\bgroup}c<{\egroup}@{}}

\newcommand{\boldgreen}[1]{\textbf{\textcolor{ForestGreen}{#1}}}
\newcommand{\boldgray}[1]{\textbf{\textcolor{gray}{#1}}}

\usepackage[accsupp]{axessibility}  

\usepackage{hyperref}
\hypersetup{
    colorlinks=true,
    linkcolor=RubineRed,
    filecolor=magenta,      
    urlcolor=RubineRed,
    pdftitle={Cross-AUC},
    pdfpagemode=FullScreen,
}

\usepackage{orcidlink}

\begin{document}

\title{When AUC Misleads: Polarization-Aware Evaluation of Deepfake Detectors under Domain Shift} 

\titlerunning{Polarization-Aware Evaluation of Deepfake Detectors under Domain Shift}

\author{Dat NGUYEN\inst{1} \and
Cosmin RADOI\inst{1} \and
Romain HERMARY\inst{1} \and 
Marcella ASTRID\inst{1} \and
Nesryne MEJRI\inst{1} \and
Enjie GHORBEL\inst{1,2} \and
Djamila AOUADA\inst{1}
}

\authorrunning{D. NGUYEN et al.}

\institute{CVI$^2$, SnT, University of Luxembourg \and
Cristal Laboratory, National School of Computer Sciences, University of Manouba
\email{firstname.lastname@uni.lu} \email{enjie.ghorbel@isamm.uma.tn}
\\
}

\maketitle

\vspace{-4mm}
\begin{abstract}
Recent advances in generative AI, such as diffusion models and face-swapping tools, have enabled the creation of highly realistic deepfakes, leading to real-world harms including financial fraud and non-consensual explicit content. In response, deepfake detection has become an active research area, with recent methods increasingly focusing on improving generalization to unseen manipulations. This is typically evaluated using the Area Under the ROC Curve (AUC) measured separately across multiple datasets. However, such an evaluation fails to reflect real-world scenarios where detectors face a mixture of data sources and varying artifact types. To address this limitation, we introduce a novel metric, Cross-dataset AUC (Cross-AUC) that averages per-domain AUCs with a measure of prediction polarization for taking into account the robustness to domain shift. The polarization extent is quantified by the Wasserstein Distance between class score distributions. Cross-AUC not only assesses the generalization capabilities of deepfake detectors under domain shifts more realistically, but it is also interpretable as it better explains the reason behind a drop in performance. Experiments performed on seven benchmark datasets demonstrate its practical relevance.

    \vspace{-2mm}
\end{abstract}

\section{Introduction}
\label{sec:intro}

Recent advances in generative AI, such as diffusion models and open-source face-swapping tools, have facilitated the creation of realistic forged facial images. Such manipulated data, also known as deepfakes, has already caused real harm, ranging from financial scams~\cite{25m_video_call} to the diffusion of non-consensual content~\cite{korea_deepfake_porn}.

\begin{figure}[t]
\centering
\includegraphics[width=0.7\linewidth]{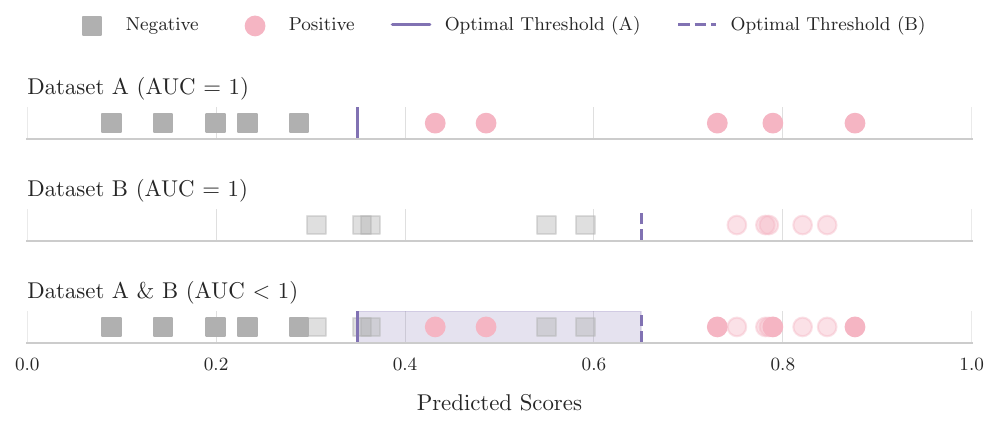}
\vspace{-3mm}
\caption{Example of optimal thresholds for predictions on different datasets, A (top) and B (middle). While a perfect $\textrm{AUC} = 1$ is reached on both datasets separately, an obvious drop in AUC will appear when combining them to simulate real-world settings (bottom), since the scores are no longer well ordered. The different nature of score distributions leads to a clear disparity between optimal thresholds (\textcolor{Orchid}{purple} range), demonstrating the importance of taking into account the polarization of probability predictions in real-world scenarios.}
\vspace{-6mm}
\label{fig:auc_weakness_vis}
\end{figure}

In response to this threat, the field of deepfake detection has become a very active research topic~\cite{ff++, ict, ete_recons, sladd, sbi, fxray, aunet, laa_net, fakeformer, fakestormer, caddm, multi-attentional, forensicsadapter, fairnessDF}. Earlier methods mostly employ binary classifiers using standard deep architectures (\eg, XceptionNet~\cite{xception}, EfficientNet~\cite{efn_net}) that learn to distinguish between real and fake data.  Nevertheless, these approaches tend to overfit specific artifact patterns introduced by the generation techniques used to create the fake training samples; thereby, achieving poor generalization capabilities when considering unseen generation techniques. Solving this issue is crucial because new generative methods are constantly emerging, introducing distinct and inherently different artifact traces. To overcome this problem, recent state-of-the-art methods have adopted multi-task learning~\cite{cstency_learning, sladd, ete_recons, laa_net, caddm} and/or data synthesis strategies~\cite{sbi, sladd, fxray, ost, plug_play}, claiming increased robustness to unseen manipulations. To demonstrate the generalization capabilities of these methods, a cross-dataset evaluation protocol is typically followed, where different datasets are employed during training and testing phases~\cite{sbi, sladd, fxray, laa_net}.
Given its robustness to imbalanced data, the Area Under the Receiver Operating Characteristic (ROC) Curve (AUC), which estimates the class separability of a binary classification model, is commonly reported as a primary evaluation metric~\cite{ict,ete_recons,sladd,sbi,fxray,aunet,laa_net,fakestormer,caddm,forensicsadapter}.
It is computed separately on each dataset and then averaged to assess the overall performance of deepfake detectors. 
Such an evaluation protocol, however, cannot fully reflect the relevance of a given deepfake detector in a real-world setting. Indeed, it assumes homogeneous score distributions across different data sources, ignoring the existence of domain shifts.
In practice, an effective detector should be able to deal with uncontrolled deepfake data featuring different kinds of artifacts at the same time, regardless of their origin.
As such, the encountered data in realistic scenarios likely resemble a mixture of multiple heterogeneous datasets.
Therefore, evaluating AUC separately on each dataset, as commonly done, fails to capture this realistic setting and tends to be overoptimistic when simply averaged.
As illustrated in Figure~\ref{fig:auc_weakness_vis}, the AUC metric can be high when evaluated on each dataset individually, while dropping drastically when considering two datasets incorporating a domain shift.

Therefore, we propose a new metric, called \textit{Cross-dataset AUC (Cross-AUC)}, that can better assess the generalization capabilities of deep detectors under domain shift.
It relies on the mean AUC calculated on diverse datasets while taking into account the polarization extent of the probability predictions.
A stronger polarization of the predicted probabilities (\ie, values closer to the boundaries $0$ and $1$) indicates a clearer separation between classes in the score space.
In such cases, a fixed threshold is more likely to remain effective under varied conditions, thereby contributing to enhanced generalization capabilities.
The polarization magnitude is estimated using the Wasserstein Distance (WD) between the distributions of positive and negative probability predictions.
Consequently, the proposed Cross-AUC captures both the relative ranking performance (via the mean AUC) and the separability extent of predictions (via the polarization), offering a more adequate estimate of generalization performance in real-world scenarios.
Finally, we empirically show that the Cross-AUC is very close the AUC measured on a combination of all the considered datasets, making it a practical measure for evaluating cross-dataset generalization.
While estimating the AUC on a combined dataset better reflects real-world settings with the available data, it lacks the interpretability provided by the introduced polarization.
In particular, it remains difficult to explain the reason behind a drop in performance of this measure compared to the average AUC computed on isolated datasets.
Therefore, Cross-AUC offers a better interpretation and understanding of the discrepancy between these two measures, demonstrating the important role of prediction polarization in the generalization capabilities of deepfake detectors.
In other words, our work suggests that the separability extent between real and fake predictions across domains is crucial.
This novel metric not only motivates the introduction of a new evaluation protocol in deepfake detection, but it also encourages researchers to develop robust deepfake detectors that consider prediction polarization as an additional learning criterion, besides traditional classification losses.

In summary, this work presents the following contributions:
\begin{itemize}
    \item We discuss and analyze the sensitivity of AUC under domain shift and its unsuitability for assessing the generalization capabilities of deepfake detectors.
    \item We rigorously define the notions of polar sets and polarity in the score space that are used to quantify the extent of polarization.
     \item We introduce a new metric called \textit{Cross-dataset AUC (Cross-AUC)} to evaluate the generalization performance of deepfake detectors, taking into account the mean AUC across different datasets, as well as the polarization of probability predictions. 
    \item We investigate the impact of prediction polarization between classes and show that similar AUC values can correspond to vastly different decision behaviors. This is quantified using the WD between score distributions.
    \item We conduct an experimental evaluation and analysis based on the proposed Cross-AUC for comparing several deepfake detectors, including recent ones using seven well-known benchmarks~\cite{ff++, celeb_df, dfd, wdf, dfdcp, dfdc, DF40}.
    
\end{itemize}

\section{Problem Formulation: AUC under Domain Shift}
\label{sec:prob_form}
\vspace{-1mm}
Let us denote by $\mathbf X \in \mathbb{R}^{c \times h\times w}$ an input image and $ y \in \{ 0,1 \}$ its associated label, which follow a joint probability distribution $p(\mathbf X,y)$. We assume that $y=0$ and $y=1$ if $\mathbf X$ belongs to the negative and positive classes (\ie, real and fake), respectively. As such, we denote by $p^p(\mathbf X )= p(\mathbf X \mid y=1) $ and $p^n(\mathbf X )= p(\mathbf X \mid y=0)$ the conditional probability density of the positive and negative classes, respectively.
Furthermore, let $s: \mathbb{R}^{c \times h\times w} \mapsto [0, 1 ] $ be a scoring function (often computed using a deep neural network in the context of deepfake detection) that estimates the positive posterior probability of an input sample $p(y=1 \mid \mathbf X)$.
Given a fixed threshold $\tau \in \mathbb{R}$, the classifier that gives the predicted label  $\hat y$  is then defined as follows: $ \hat y=  \mathbb{I}(p(y=1 \mid \mathbf X)>\tau)$, where $\mathbb{I}$ is the indicator function.
The Area Under the ROC Curve (AUC) is a standard metric often used in deepfake detection and more generally in binary classification. Specifically, it reflects how well a model is able to separate positive from negative samples in the score space and is independent of the classification threshold $\tau$. It measures the probability that a randomly chosen positive instance (\eg, a fake video) receives a higher confidence score than a randomly chosen negative one (\eg, a real video)~\cite{AUC_maxi,auc_acc_evaluation,partial_auc_maxi,misuse_AUC}. Formally, the AUC can be defined as,
\begin{equation}
\label{equa:auc_overall}
\text{AUC}(s) = \mathbb{E}_{\mathbf{X}^p \sim p^p(\mathbf{X}),\, \mathbf{X}^n \sim p^n(\mathbf{X})} \left[ \mathbb{I}(s(\mathbf{X}^p) > s(\mathbf{X}^n)) \right] \text{,}
\end{equation}
This formulation emphasizes that AUC reflects the ability of a model to rank positive instances higher than negative ones, depending only on the relative ordering of the scores.

Although AUC is effective from a dataset to another when the class-conditional distributions are stable, it becomes problematic in the presence of a domain shift. Let $i$ and $j$ denote two different datasets with their distribution of positive and negative prediction samples denoted as $p_i^p(\mathbf X )$, $p_j^p(\mathbf X )$ and $p_i^n(\mathbf X )$, $p_j^n(\mathbf X )$, respectively.

In the presence of a \textbf{complete domain overlap} between $i$ and $j$, the class-conditional distributions are approximately aligned:
\begin{equation}
p_i^p(\mathbf X ) \approx p_j^p(\mathbf X ), \quad p_i^n(\mathbf X ) \approx p_j^n(\mathbf X ) \text{.}
\end{equation}
Hence, the model produces comparable confidence scores across domains. The indicator function in the AUC formula, \( \mathbb{I}(s(\mathbf{x}^P) > s(\mathbf{x}^N)) \), remains meaningful, and global AUC reliably reflects the model's performance.

Nevertheless, in the presence of a \textbf{domain gap} in the prediction probability space, the two distributions differ significantly,
\begin{equation}
p_i^p(\mathbf X ) \not\approx p_j^p(\mathbf X ),  \quad p_i^n(\mathbf X )\not\approx p_j^n(\mathbf X )  \text{.}
\end{equation}

In this context, the positive and negative posterior probabilities $p^p(\mathbf  X)$ and $p^n(\mathbf  X)$ can be expressed as a mixture of dataset-specific posterior probabilities as follows,
\begin{align}
p^p(\mathbf  X) = \sum_i \alpha_i \, p^p_i(\mathbf X ) \not\approx  p^p_i(\mathbf X ), \text{ } \forall i \text{,} \quad 
p^n(\mathbf X) = \sum_i \beta_i \, p^n_i(\mathbf X ) \not\approx  p^n_i(\mathbf X ),  \text{ } \forall i \text{,} 
\end{align}
where $\alpha_i$ and $\beta_i$  represent the  mixture weights.  
This is typically the case when considering different deepfake detection datasets such as  FF++~\cite{ff++}, CDF~\cite{celeb_df}, DFDC~\cite{dfdc}, and DFW~\cite{wdf}.
This divergence can result from differences in deepfake generation methods, compression levels, facial attributes, resolution, or lighting conditions across datasets. This can lead to inconsistencies in how the model scores samples across domains, as we can deduce from Eq.~\eqref{equa:auc_overall}. For example, the model might assign higher confidence scores to fake videos from one dataset, but not to the other dataset. These inconsistencies can distort global rankings and degrade AUC, even when performance within each domain is strong. This makes AUC an unreliable performance measure if considered on different domains separately, as experimentally demonstrated in Section~\ref{sec:results_and_discussions}.

\begin{figure}[t]
  \centering
  \begin{minipage}{0.49\linewidth}
    \centering
    \includegraphics[width=0.8\linewidth]{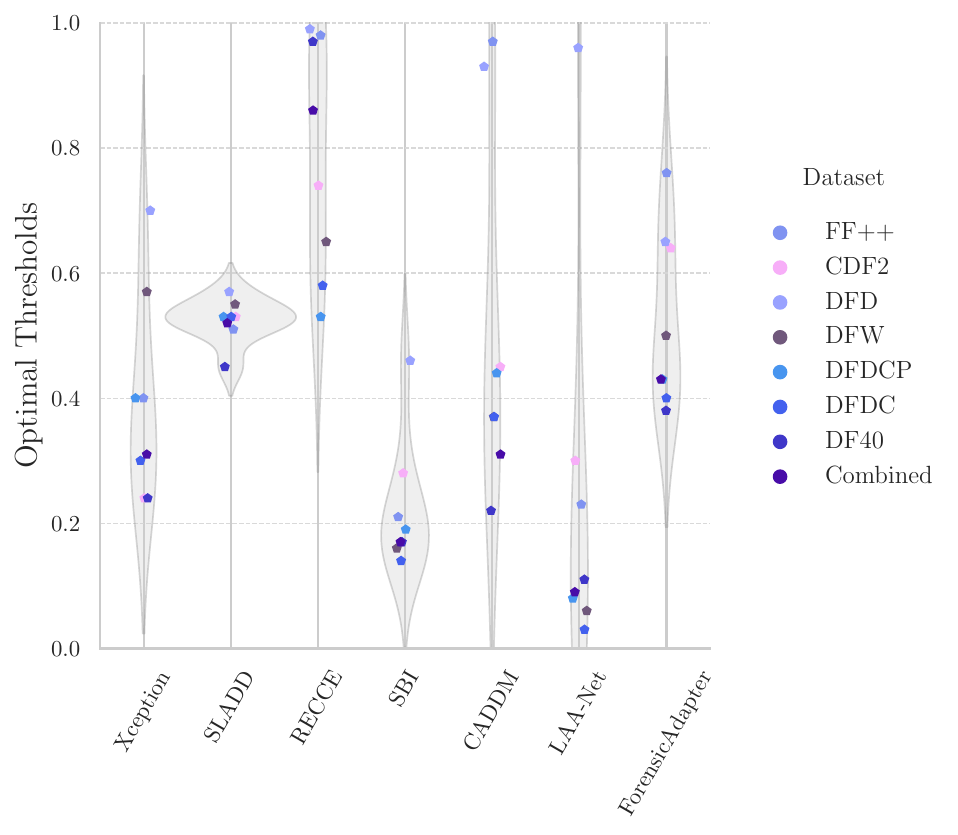}
    \vspace{-4mm}
    \caption{Optimal thresholds for the studied SOTA methods on the seven different datasets~\cite{ff++, celeb_df, dfd, wdf, dfdcp, dfdc, DF40} (and all of them combined).}
    \label{fig:thresh_per_model}
  \end{minipage}
  \hfill
  \begin{minipage}{0.49\linewidth}
    \centering
    \includegraphics[width=0.8\linewidth]{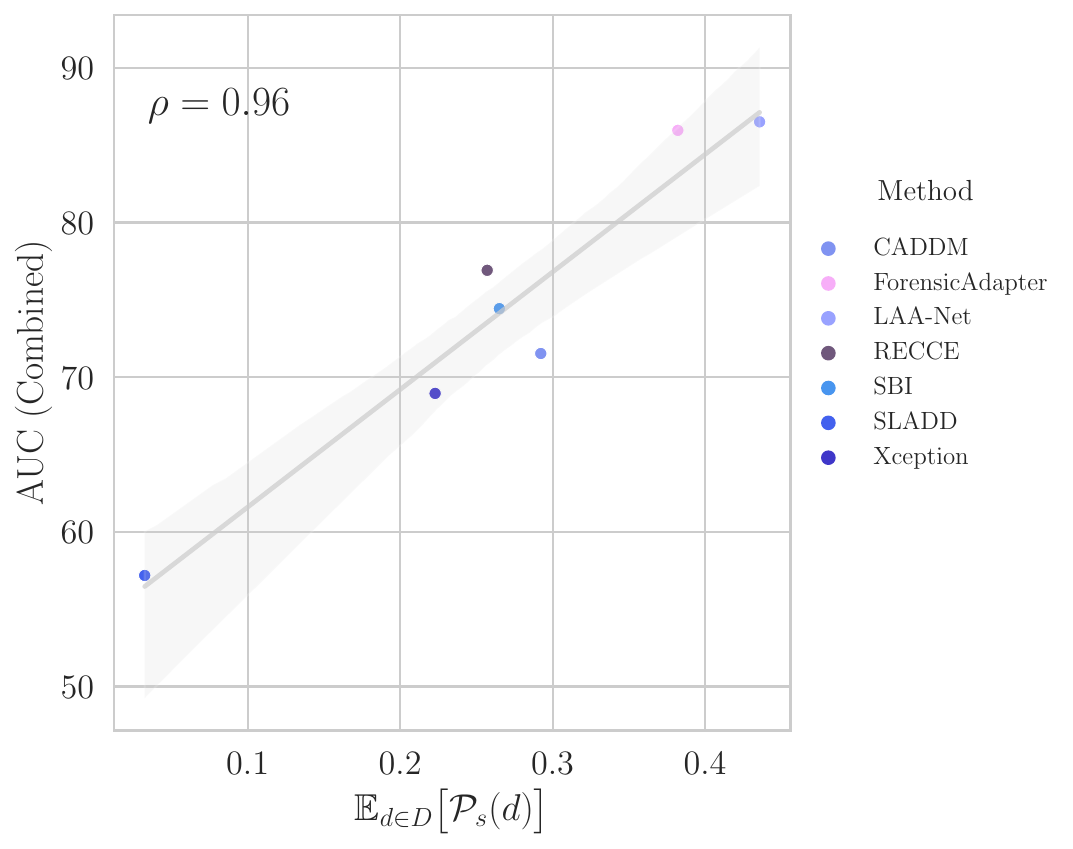}\\[1.4em]
    \vspace{-7mm}
    \caption{Correlation between estimated polarity (averaged across datasets) and the target combined AUC. Pearson coefficient $\rho = 0.96$.}
    \label{fig:corr_distance}
  \end{minipage}
  \vspace{-5mm}
\end{figure}

\section{Cross-Dataset AUC}
\label{sec:method}

We propose a revised evaluation protocol that more faithfully reflects the deployment challenges of deepfake detection.
As discussed in Section~\ref{sec:prob_form}, the conventional evaluation metric, AUC, is computed independently on individual datasets and provides an incomplete or overly optimistic view of model robustness.
This practice overlooks two critical issues: \textit{(i) the instability of decision thresholds}, and \textit{(ii) the implicit assumption of the non-existence of distribution shifts}, across datasets.
In this section, we propose to define the notion of polarization in the score space, which is used as a new measure for assessing threshold instability. The proposed polarization measure is then integrated into the newly introduced metric termed Cross-AUC, which is proposed as a replacement for the standard mean AUC under domain shift.

\subsection{Threshold Instability and Polarization}

We first expose score instabilities in deepfake detection models.
We verify it empirically by locating the separation between the negative and positive probability predictions for a given method over multiple datasets.
This separation is known as a decision threshold and is necessary for any deployment scenario.
Its value can depend on the application (\eg, maximizing detection rate or reducing false alarm rate), but in a generic study, an ideal threshold should be the best trade-off between correctly classified positives and negatives.

Given a scoring function $s$ and a dataset $d$, we denote the set of obtained scores on $d$ using the function $s$ as $\mathcal{O}_{s,d} = \{s(\mathbf{X}), \forall\mathbf{X}\in d\}$.

We choose the corresponding optimal threshold $\tau_{s,d}$ that gives the closest score to perfect classification (i.e, $\text{FPR} = 0$, $\textrm{TPR} = 1$).

From the set of all possible thresholds $\mathcal{T}$, it is obtained by minimizing the distance between the FPR and TPR values from the ROC curve and the FPR and TPR obtained under perfect classification.
\begin{equation}
    \tau_{s,d} = \argmin_{t \in \mathcal{T}}\left(\sqrt{\left[1 - \text{TPR}_t(\mathcal{O}_{s,d})\right]^2 + \text{FPR}_t(\mathcal{O}_{s,d})^2}\right).
\end{equation}

In Figure~\ref{fig:thresh_per_model}, we show the optimal thresholds found for the considered SOTA methods across several datasets.
Our analysis reveals noticeable threshold variations, with standard deviations reaching $0.16$ for Xception~\cite{ff++} and even $0.31$ for LAA-Net~\cite{laa_net}, demonstrating the impact of shifts between testing data sources.
This large discrepancy between optimal thresholds will cause lower performance than expected after deployment.

\begin{figure*}[t]
\centering
\includegraphics[width=0.9\linewidth]{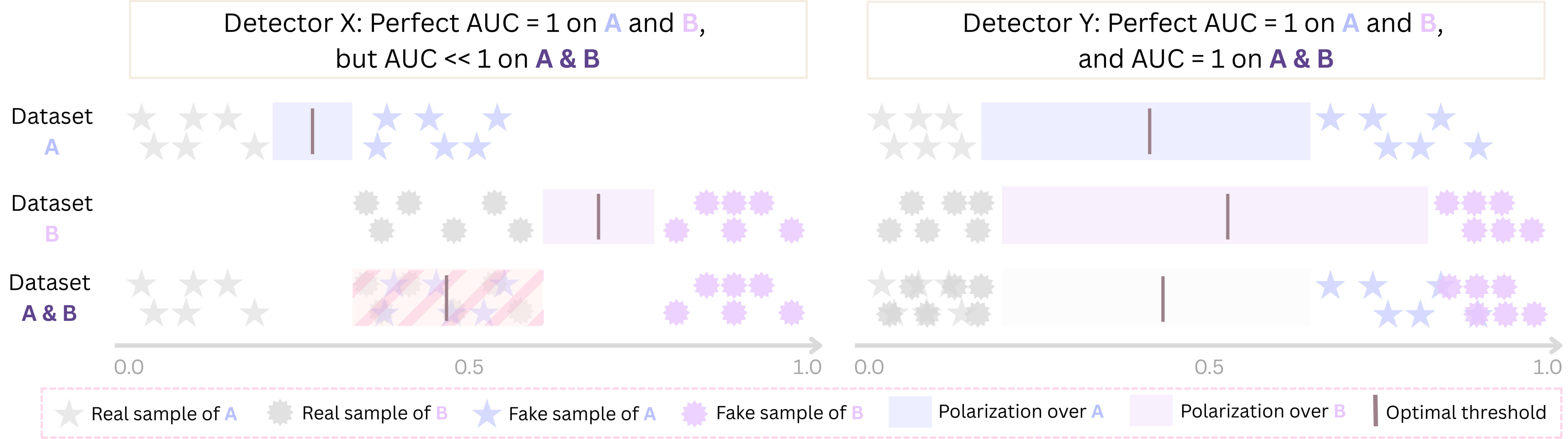}
\vspace{-1mm}
\caption{Example of the obtained optimal thresholds on two datasets, A (top) and B (middle). Although both detectors achieve perfect $\text{AUC}=1$ when evaluated on each dataset separately, mixing A\&B (bottom) can cause a sharp AUC drop (Detector X), while a more stable detector preserves ordering and maintains $\text{AUC}=1$ (Detector Y). It can be observed that a larger gap exists between the two classes in the score space, reducing the impact of a varying threshold. This motivates adopting a polarization-aware evaluation.}
\label{fig:auc_weakness_vis_method}
\vspace{-5mm}
\end{figure*}

Although the instability of the optimal threshold across datasets reflects the instability of a deepfake detector model, its stability does not guarantee strong generalization capabilities.
In particular, if the predicted probabilities vary only within a restricted range, the variance of optimal thresholds might be small. Yet, the corresponding model can still be unstable across datasets with disruptive domain shifts in the score space.
On the other hand, a model that is robust to high variance thresholds is more likely to be stable across domain shifts. This intuition is illustrated in Figure~\ref{fig:auc_weakness_vis_method}. 

Therefore, we propose measuring this robustness criterion via \textit{polarization}.
Herein, we start by introducing \textbf{\textit{Polar Sets}} and \textbf{\textit{Polarity}} as follows.

\textit{\textbf{Definition (Polar Sets)}}: Let $s \in \mathcal S$ be a scoring function (\eg, a model) and $d \in \mathcal D$ a dataset. The polar sets associated to $s$ and $d$ are defined as $\mathcal R_{s,d} = \{s(\mathbf{X}) \mid y = 0, \forall \mathbf X \in d\}$ and $\mathcal F_{s,d} = \{s(\mathbf{X}) \mid y = 1, \forall \mathbf{X} \in d\}$.
Thus, the sets $\mathcal R_{s,d}$ and $\mathcal F_{s,d}$ contain the probability predictions given by $s$ of all real data and fake data, respectively. 

\textit{\textbf{Definition (Polarity)}}: The polarity $P_s(d)$ of a scoring function $s \in \mathcal S$ on a dataset $d \in\mathcal D$ is a measure of divergence (or distance) between the estimated probability distributions $R_{s,d} = \Theta(\mathcal R_{s,d})$ and ${F}_{s,d} = \Theta(\mathcal F_{s,d})$, with $\Theta$ a given density estimation function.
We refer to the measuring function as $\Pi$, such that $P_s(d) = \Pi(R_{s,d}, F_{s,d})$.

An ideal model $s^*$ would converge towards ${R}_{s,d} = \delta_0$ and ${F}_{s,d} = \delta_1$, which represent Dirac impulses at both extremities of the scoring space $[0,1]$.
By assuming that the distance between the distribution is calculated using the Wasserstein distance ($\text{WD}$), which we will use throughout the whole study, we can intuitively find that $s^*$ has a polarity ${P}_{s*,d}$~~of $\text{WD}(\delta_0, \delta_1) = |0 - 1| = 1$, for any $d \in \mathcal D$. On the other hand, the worst performing model $s^-$ would end up with ${R}_{s^-,d} \approx {F}_{s^-,d}$ and a polarity ${P}_{s^-, d}\approx 0$.

As we focus on the robustness of models approaching perfect convergence, we can safely assume both
densities ${R}_{s,d}$ and ${F}_{s,d}$ to be close to the objective densities (${R}_{s,d} \approx\delta_\alpha$, ${F}_{s,d} \approx \delta_{1-\beta}$, $\alpha$ and $\beta$ representing small shifts).
Upon this condition, we can derive the definition space of $\tau_{s,d}$ for each dataset $d \in \mathcal{D}$, denoted as $[\alpha_{s,d}, 1 - \beta_{s,d}]$ and deduce a tolerance of $1 - \underset{d \in D}{\max}(\beta_{s,d}) - \underset{d \in D}{\max}(\alpha_{s,d})$ for the cross-dataset optimal threshold $\tau_s$ of the model $s$.
This suggests that the lower $\alpha$ and $\beta$ are across numerous datasets, the greater the tolerance over different thresholds will be. This directly translates to having a maximum polarity across datasets.
The hypothesis is empirically verified and illustrated in Figure~\ref{fig:corr_distance}, where we visualize  the correlation between individual dataset polarities and the AUC on a \textbf{Combined dataset} obtained by aggregating all the datasets. This allows simulating real-world settings where data are unconstrained, thereby assessing more faithfully the deployment readiness. Specifically, the AUC results on the Combined dataset are reported according to the empirical average of polarities, $\mathbb{E}_{d \in D}\bigl[{P}_s(d)\bigr]$. In summary, a high correlation between the empirical average of polarities, $\mathbb{E}_{d \in D}\bigl[{P}_s(d)\bigr]$ and the AUC can be observed.

\subsection{Leveraging Polarization and AUC for Cross-Dataset Evaluation}
\label{sec:new metric}

To better evaluate the reliability of a model and take into account conflicting optimal thresholds, we propose to use the defined polarity as a reward parameter.
We further establish that a stable AUC and polarities throughout the datasets will ensure better reliability.
To this end, we define a penalty-reward criterion ${C}$, such that we have \textit{i) a penalty} when AUCs are unstable and \textit{ii) a reward} for high polarities.

For a model $s \in \mathcal S$, we will thus define the criterion over its corresponding sets of AUC scores $\mathcal A = \{\text{AUC}(s, d),\,\forall d \in \mathcal D\}$ and polarities $\mathcal P = \{{P}_s(d),\,\forall d \in \mathcal D\}$.
To penalize any instability of AUCs and polarities, we quantify the variability of the given sets with a measure of central dispersion (\eg, standard deviation), which we denote as ${\Phi}: \mathbb{R}_+^k \mapsto \mathbb{R}_+$ (where $k = \text{card}(\mathcal D)$ is the number of datasets).
Taking the opposite of this quantity, $-\Phi(\cdot)$, yields the desired penalties.
To reward the score polarization, we rely on the estimation of the global model behavior with a measure of central tendency (\eg, harmonic mean), which we denote as ${\Psi}: \mathbb{R}_+^k \mapsto \mathbb{R}_+$.
We define the criterion $C$ by compiling both sub-criteria as follows:
%
\begin{equation}
    {C}(\mathcal A, \mathcal P) = - \Phi(\mathcal A) + \left| {\Psi}(\mathcal P) - {\Phi}(\mathcal P) \right|.
\end{equation}
%
We apply the same estimator $\Psi$ over the AUCs to obtain the performance tendency of $s$, and combine it with the previously defined criterion to form the proposed Cross-AUC:
\begin{equation}
\label{eq:cross_auc}
    \text{Cross-AUC} ={\Psi}(\mathcal A) +  \lambda \cdot {C}(\mathcal A,\mathcal P),
\end{equation}
with $\lambda$ a balancing parameter.
We provide a comprehensive analysis of the selection of Cross-AUC components in the next section.

\section{Experiments}
\label{sec:ex}
\vspace{-1mm}

In this section, we experimentally demonstrate that the proposed Cross-AUC is a reliable estimator of model stability and provides values that are very close to the AUC on the Combined dataset.
We first present the experimental settings in Section~\ref{subsec:settings} and discuss the obtained results in Section~\ref{sec:results_and_discussions}.

\vspace{-2mm}
\subsection{Experimental Settings}
\label{subsec:settings}
\vspace{-1mm}

\noindent\textbf{Deepfake Detectors.} To conduct our evaluation study, seven SOTA methods~\cite{ff++, sladd, ete_recons, sbi, caddm, laa_net, forensicsadapter} are selected. The selection is based on the following criteria: (1) the official codes of the considered methods are available and open-source for reliable reproducibility; (2) they are diverse in the way they approach deepfake detection: we include in our study an end-to-end binary classifier (\textbf{Xception}~\cite{ff++} ), a disentanglement learning approach (\textbf{RECCE}~\cite{ete_recons}), a method relying on data synthesis (\textbf{SBI}~\cite{sbi}), two hybrid methods combining multi-task learning and data synthesis strategies (\textbf{SLADD}~\cite{sladd}, \textbf{CADDM}~\cite{caddm}), a fine-grained approach (\textbf{LAA-Net}~\cite{laa_net}), and a CLIP-based~\cite{clip} adapter method (\textbf{ForensicAdapter}~\cite{forensicsadapter}); (3) the considered baselines are recent and have been accepted in top-tier conferences (at the exception of ~\cite{ff++} that was published in $2019$).
The summary of the compared deepfake detectors is presented in Table~\ref{tab:deepfake_detectors}.
More details are provided in the supplementary materials.

\begin{table*}
   \centering
   \vspace{-6mm}
   \caption{Summary of compared deepfake detectors.}
   \vspace{-3mm}
   \resizebox{\linewidth}{!}{
   \begin{tabular}{c|c|c|c|c|c}
    \toprule
       Detector & Backbone & Family & Respository & Weights Release & Venue\\
    \midrule
    \midrule
       Xception~\cite{ff++} & Xception~\cite{xception} & Binary Classifier & \url{https://github.com/ondyari/FaceForensics} & \checkmark & ICCV'19 \\
       SLADD~\cite{sladd} & Modified Xception~\cite{xception} & Multitask + Data synthesis & \url{https://github.com/liangchen527/SLADD} & $\times$ & CVPR'22 \\
       RECCE~\cite{ete_recons} & Designed Networks & Disentanglement Learning & \url{https://github.com/VISION-SJTU/RECCE} & \checkmark & CVPR'22 \\
       SBI~\cite{sbi} & EfficientNet~\cite{efn_net} & Data synthesis & \url{https://github.com/mapooon/SelfBlendedImages} & \checkmark & CVPR'22 \\
       CADDM~\cite{caddm} & EfficientNet~\cite{efn_net} + Multiscale & Multitask + Data synthesis & \url{https://github.com/megvii-research/CADDM} & \checkmark & CVPR'23 \\
       LAA-Net~\cite{laa_net} & EfficientNet~\cite{efn_net} + E-FPN & Multitask + Data synthesis + Fine-grained & \url{https://github.com/10Ring/LAA-Net} & \checkmark & CVPR'24 \\
        ForensicAdapter~\cite{forensicsadapter} & CLIP~\cite{clip} + ViT~\cite{ViT} & Multitask + Data synthesis + Adapting CLIP & \url{https://github.com/OUC-VAS/ForensicsAdapter} & \checkmark & CVPR'25 \\
    \bottomrule
    \end{tabular}
    }
    \label{tab:deepfake_detectors}
    \vspace{-6mm}
\end{table*}

\noindent\textbf{Datasets.}
Seven standard and challenging datasets are used in our experiments, including \textbf{FaceForensics++} (FF++)~\cite{ff++}, \textbf{Celeb-DF} (CDF)~\cite{celeb_df}, \textbf{Google Deepfake Detection} (DFD)~\cite{dfd}, \textbf{WildDeepfake} (DFW)~\cite{wdf}, \textbf{Deepfake Detection Challenge} (DFDC)~\cite{dfdc}, \textbf{Deepfake Detection Challenge Preview} (DFDCP)~\cite{dfdcp}, and \textbf{DF40}~\cite{DF40}. FF++ is typically used for training, while the others serve as test sets in the cross-dataset evaluation protocol.
These datasets are selected based on the following criteria: 1) they are widely used for evaluation in the research community; 2) they contain fakes from diverse sources, including those generated by black-box or undisclosed methods, for which no implementation details or prior knowledge are available; 3) they present increased complexity as a result of various image perturbations, such as compression, noise, or blur.
An overview of the selected benchmarks and further details is provided in the supplementary materials.

\noindent\textbf{Implementation details.} We extract $32$ frames per video following the conventional test split~\cite{ff++}. Facial regions are cropped using Face-RetinaNet~\cite{retina_face}. The bounding boxes are slightly enlarged by a factor of $1.25$ around the center of the face and then resized to a fixed resolution of $256 \times 256$. For all experiments, we use the official pre-trained models of the selected methods. Since SLADD does not provide pre-trained weights, we retrain it on FF++ following its open-source implementation. 
Unless specified differently, we use the harmonic mean for $\Psi$, the standard deviation for $\Phi$, the Quantile Cumulative Density Function ($\textrm{CDF}^{-1}$) for the density estimation function ${D}$,  and $\lambda = 0.5$ to provide the best results since they yield the best performance as demonstrated empirically in Section~\ref{sec:results_and_discussions}.

\noindent\textbf{Evaluation Protocol.}
We consider two evaluation settings.
The first is the conventional protocol, in which each model is evaluated separately on each dataset.
The resulting independent scores are used to compare the AUC with the proposed Cross-AUC metric.
To better reflect the heterogeneity and multi-source nature of real-world data, the second protocol is based on the evaluation of each model on a unified dataset, referred to as the ``\textit{Combined}'' dataset, obtained by aggregating all seven datasets.
This enables the exposure of deepfake detection methods to domain gaps that may remain hidden in single-dataset evaluation settings.

\noindent\textbf{Evaluation Metrics\label{subsec:eval_metric}.} 
The polarization is computed using the Wasserstein Distance (WD) for $\Pi$, measuring the distance between the estimated densities $R_{s,d}$ and $F_{s, d}$ (real and fake score distributions).
To approximate these densities, we experimented with different density estimation functions $\Theta$ based on non-parametric and parametric approaches.
For the former, we employ Kernel Density Estimation (KDE) and Quantile Cumulative Density Function (Q); for the latter, we consider two types of distributions, namely, Gaussian Mixture Model (GMM) and Beta Distribution (BD).
To estimate the parameters of GMM and BD, Expectation-Maximization (EM) and numerical Maximum Likelihood Estimation (MLE) are used, respectively.
\textit{By default, Q is chosen for its effectiveness}.
We mainly focus on comparing the averaged AUC (AUC$_a$), the AUC on the Combined dataset (AUC$_c$), and our Cross-AUC.
Comparison with additional evaluation metrics is provided in the supplementary materials for a more in-depth analysis, including Accuracy, Balanced Accuracy, Precision, Recall, Specificity, F1-Score, and Equal Error Rate.



\begin{table}
\centering
\caption{Performance of the selected models on all datasets. 
\textbf{$\tau$}: Optimal threshold; 
$\phi_{\tau}$: variance of the top–\(K\) optimal thresholds;
$W_\text{KDE}$, $W_\text{Q}$, $W_\text{GMM}$, $W_\text{BD}$: model polarities using WD as $\Pi$ over different density estimation ($\Theta$) methods, including non-parametric (Kernel Density Estimation (KDE), Quantile (Q)), and parametric ones assuming two types of distributions, \ie, GMM and BD.
We set $\Phi$ as the standard deviation, $\Psi$ as the harmonic mean and $\lambda=0.5$.
\textcolor{gray}{Gray} represents the datasets the model was trained on. \boldgreen{Green} highlights the best AUC and polarities.}
\scalebox{0.72}{
\begin{tabular}{ll|c|cccccccc}
\toprule
 & & \textcolor{gray}{FF++} & CDF & DFD & DFW & DFDCP & DFDC & DF40 & Average & Combined \\
 \midrule
 \midrule
 \multirow{7}{*}{Xception} & AUC (\%) & \textcolor{gray}{93.6} & 62.1 & 92.24 & 63.71 & 70.56 & 58.98 & 81.59 & 74.68 & 68.95 ($\textcolor{red}{\downarrow 5.73}$)
\\ & $\tau$ & \textcolor{gray}{0.4} & 0.24 & 0.7 & 0.57 & 0.4 & 0.3 & 0.24 & 0.4 & 0.31 
\\ & $\phi_{\tau}$ & \textcolor{gray}{0.0004} & 0.0004& 0.0002& 0.0007& 0.0010& 0.0001 & 0.0001 & 0.0004& 0.0001\\
& $W_\text{KDE}$ & \textcolor{gray}{0.524} & 0.064 & 0.430 & 0.110 & 0.146 & 0.078 & 0.239 & 0.227 & 0.168 
\\ & $W_\text{BD}$ & \textcolor{gray}{0.549} & 0.068 & 0.446 & 0.122 & 0.164 & 0.060 & 0.232 & 0.234 & 0.152
\\ & $W_\text{Q}$ & \textcolor{gray}{0.661} & 0.077 & 0.524 & 0.139 & 0.183 & 0.091 & 0.264 & 0.277 & 0.190
\\ & $W_\text{GMM}$ & \textcolor{gray}{0.616} & 0.069 & 0.496 & 0.149 & 0.118 & 0.081 & 0.249 & 0.254 & 0.180
\\
\midrule
\multirow{7}{*}{SLADD} & AUC (\%) & \textcolor{gray}{88.31} & 56.65 & 97.22 & 48.63 & 65.98 & 57.37 & 72.25 & 69.48 & 57.18 ($\textcolor{red}{\downarrow 12.3}$)
\\ & $\tau$& \textcolor{gray}{0.51} & 0.53 & 0.57 & 0.55 & 0.53 & 0.53 & 0.45 & 0.52 & 0.52 
\\ & $\phi_{\tau}$ & \textcolor{gray}{0.0001} & 0.0001& 0.0001 & 0.0001& 0.0001& 0.0001 & 0.0001 & 0.0001& 0.0001
\\ & $W_\text{KDE}$& \textcolor{gray}{0.060} & 0.001 & 0.098 & 0.004 & 0.022 & 0.012 & 0.003 & 0.028 & 0.013 
\\ & $W_\text{BD}$& \textcolor{gray}{0.078} & 0.003 & 0.079 & 0.001 & 0.007 & 0.008 & 0.024 & 0.029 & 0.013 
\\ & $W_\text{Q}$& \textcolor{gray}{0.079} & 0.005 & 0.090 & 0.002 & 0.015 & 0.010 & 0.026 & 0.032 & 0.014
\\ & $W_\text{GMM}$& \textcolor{gray}{0.079} & 0.005 & 0.089 & 0.004 & 0.015 & 0.010 & 0.027 & 0.033 & 0.014
\\ 
\midrule
\multirow{7}{*}{RECCE} & AUC (\%) & \textbf{\textcolor{gray}{99.56}} & 69.02 & 97.82 & 65.19 & 82.24 & 71.19 & 82.94 & 81.13 & 76.91 ($\textcolor{red}{\downarrow 4.22}$)
\\ & $\tau$& \textcolor{gray}{0.98} & 0.74 & 0.99 & 0.65 & 0.53 & 0.58 & 0.97 & 0.77 & 0.86 
\\ & $\phi_{\tau}$ & \textcolor{gray}{0.00001} & 0.0002& 0.0001& 0.0001& 0.0001& 0.0001 & 0.0001& 0.0001& 0.0001
\\ & $W_\text{KDE}$& \textcolor{gray}{0.631} & 0.126 & 0.352 & 0.096 & 0.212 & 0.181 & 0.153 & 0.250 & 0.221 
\\ & $W_\text{BD}$& \textcolor{gray}{0.569} & 0.124 & 0.280 & 0.116 & 0.240 & 0.182 & 0.117 & 0.233 & 0.201 
\\ & $W_\text{Q}$& \textcolor{gray}{0.716} & 0.136 & 0.355 & 0.115 & 0.243 & 0.203 & 0.144 & 0.273 & 0.232
\\ & $W_\text{GMM}$& \textcolor{gray}{0.692} & 0.134 & 0.349 & 0.111 & 0.227 & 0.203 & 0.142 & 0.265 & 0.228 
\\ 
\midrule
\multirow{7}{*}{SBI} & AUC (\%) & \textcolor{gray}{98.23} & 84.65 & 96.78 & 60.83 & 86.83 & 69.77 & 72.05 & 81.3 & 74.43 ($\textcolor{red}{\downarrow 6.87}$)
\\ & $\tau$& \textcolor{gray}{0.21} & 0.28 & 0.46 & 0.16 & 0.19 & 0.14 & 0.17 & 0.23 & 0.17 
\\ & $\phi_{\tau}$ & \textcolor{gray}{0.0006} & 0.0001& 0.0008& 0.0001& 0.0002& 0.0001 & 0.0001& 0.0002& 0.0001
\\ & $W_\text{KDW}$& \textcolor{gray}{0.503} & 0.215 & 0.516 & 0.078 & 0.220 & 0.097 & 0.103 & 0.248 & 0.152 
\\ & $W_\text{BD}$& \textcolor{gray}{0.513} & 0.219 & 0.500 & 0.089 & 0.224 & 0.116 & 0.112 & 0.253 & 0.182 
\\ & $W_\text{W}$& \textcolor{gray}{0.576} & 0.238 & 0.577 & 0.085 & 0.244 & 0.108 & 0.116 & 0.278 & 0.172 
\\ & $W_\text{GMM}$& \textcolor{gray}{0.567} & 0.236 & 0.564 & 0.084 & 0.245 & 0.105 & 0.115 & 0.274 & 0.171
\\ 
\midrule
\multirow{7}{*}{CADDM}& AUC (\%) & \textcolor{gray}{99.26} & 80.7 & 99.52 & 76.31 & 71 & 70.33 & 76.28 & 81.91 & 71.53 ($\textcolor{red}{\downarrow 10.38}$)
\\ & $\tau$& \textcolor{gray}{0.97} & 0.45 & 0.93 & 0.37 & 0.44 & 0.37 & 0.22 & 0.53 & 0.31 
\\ & $\phi_{\tau}$ & \textcolor{gray}{0.0253} & 0.0001& 0.0001& 0.0001& 0.0011& 0.0001 & 0.0001& 0.0038& 0.0001
\\ & $W_\text{KDE}$& \textcolor{gray}{0.661} & 0.203 & 0.436 & 0.210 & 0.156 & 0.155 & 0.137 & 0.280 & 0.176 
\\ & $W_\text{BD}$& \textcolor{gray}{0.607} & 0.203 & 0.386 & 0.238 & 0.182 & 0.165 & 0.146 & 0.275 & 0.214 
\\ & $W_\text{Q}$& \textcolor{gray}{0.726} & 0.226 & 0.466 & 0.246 & 0.179 & 0.174 & 0.151 & 0.310 & 0.198 
\\ & $W_\text{GMM}$& \textcolor{gray}{0.714} & 0.224 & 0.459 & 0.241 & 0.174 & 0.173 & 0.150 & 0.305 & 0.197 
\\ 
\midrule
\multirow{7}{*}{LAA-Net} & AUC (\%) & \textcolor{gray}{99.45} & \boldgreen{95.45} & 98.47 & 79.52 & 86.48 & 72.6 & 86.04 & 88.28 & \boldgreen{86.51} ($\textcolor{red}{\downarrow 1.77}$)
\\ & $\tau$ & \textcolor{gray}{0.23} & 0.3 & 0.96 & 0.06 & 0.08 & 0.03 & 0.11 & 0.25 & 0.083 
\\ & $\phi_{\tau}$ & \textcolor{gray}{0.0161} & 0.0004& 0.0001& 0.0001& 0.0005& 0.0001& 0.0001& 0.0024& 0.0001\\ 
& $W_\text{KDE}$& \textbf{\textcolor{gray}{0.831}} & \boldgreen{0.511} & 0.633 & \boldgreen{0.262} & \boldgreen{0.334} & 0.169 & 0.281 & \boldgreen{0.431} & 0.302 
\\ 
& $W_\text{BD}$& \textbf{\textcolor{gray}{0.749}} & \boldgreen{0.495} & 0.547 & 0.248 & \boldgreen{0.357} & 0.189 & 0.272 & \boldgreen{0.408} & 0.296 
\\ 
& $W_\text{Q}$& \textbf{\textcolor{gray}{0.927}} & \boldgreen{0.582} & 0.708 & \boldgreen{0.285} & \boldgreen{0.446} & 0.156 & 0.299 & \boldgreen{0.486} & 0.318 
\\ 
& $W_\text{GMM}$& \textbf{\textcolor{gray}{0.911}} & \boldgreen{0.579} & 0.690 & \boldgreen{0.292} & \boldgreen{0.417} & 0.154 & 0.286 & \boldgreen{0.475} & 0.317 
\\
\midrule
\multirow{7}{*}{ForensicAdapter} & AUC (\%) & \textcolor{gray}{98.39} & 94.94 & \boldgreen{99.8} & \boldgreen{80.95} & \boldgreen{87.04} & \boldgreen{83.88} & \boldgreen{86.3} & \boldgreen{90.18} & 85.96 ($\textcolor{red}{\downarrow 4.22}$)
\\ & $\tau$& \textcolor{gray}{0.76} & 0.64 & 0.65 & 0.5 & 0.43 & 0.4 & 0.38 & 0.53 & 0.43
\\ & $\phi_{\tau}$ & \textcolor{gray}{0.0013} & 0.0003& 0.0027& 0.0001& 0.0007& 0.0001& 0.0001& 0.0007& 0.0001
\\ & $W_\text{KDE}$& \textcolor{gray}{0.497} & 0.338 & \boldgreen{0.659} & 0.251 & 0.259 & \boldgreen{0.257} & \boldgreen{0.291} & 0.365 & \boldgreen{0.304}
\\ & $W_\text{BD}$& \textcolor{gray}{0.460} & 0.334 & \boldgreen{0.624} & \boldgreen{0.253} & 0.268 & \boldgreen{0.266} & \boldgreen{0.294} & 0.357 & \boldgreen{0.306}
\\ & $W_\text{Q}$& \textcolor{gray}{0.540} & 0.374 & \boldgreen{0.713} & 0.285 & 0.293 & \boldgreen{0.286} & \boldgreen{0.322} & 0.402 & \boldgreen{0.336}
\\ & $W_\text{GMM}$& \textcolor{gray}{0.521} & 0.371 & \boldgreen{0.709} & 0.284 & 0.288 & \boldgreen{0.285} & \boldgreen{0.320} & 0.397 & \boldgreen{0.335} \\
\bottomrule
\end{tabular}
}
\label{tabl:trimmed_results}
\end{table}

\vspace{-2mm}
\subsection{Results and Discussion}\label{sec:results_and_discussions}

In this section, we present and discuss the obtained results. Additional results are provided in the supplementary materials, such as the runtime analysis, the effect of the number of domains, and the performance obtained using additional metrics.

\noindent\textbf{Overall Performance Analysis.}
Table~\ref{tabl:trimmed_results} summarizes per-dataset AUC, average AUC, AUC on a combination of seven considered datasets~\cite{ff++, celeb_df, dfd, wdf, dfdcp, dfdc, DF40}, optimal thresholds ($\tau$), and their stability ($\phi_{\tau}$), as well as score polarization (polarity) measured via WD using different density estimators. 
As shown, ForensicAdapter~\cite{forensicsadapter} and LAA-Net~\cite{laa_net} emerge as the strongest performers under domain shift. ForensicAdapter attains the highest mean AUC ($90.18\%$) while maintaining a high combined AUC ($85.96\%$), and it consistently yields strong polarity, indicating well-separated score distributions that transfer across datasets. LAA-Net achieves the best combined AUC ($86.51\%$) and leads in terms of polarity on most datasets, suggesting a strong discriminative capacity. However, its higher threshold variance $\phi_{\tau}$ suggests a less stable decision boundary in cross-domain settings. RECCE~\cite{ete_recons} achieves consistent performance across domains with stable AUC ($81.13\%$) and Combined AUC ($76.91\%$), and low threshold variance. Nonetheless, its class separability capacity is slightly weaker as compared to LAA-Net and ForensicAdapter.
SBI~\cite{sbi} and CADDM~\cite{caddm} perform moderately well in terms of both average and Combined AUC, but their optimal threshold vary importantly.
Xception~\cite{xception} and SLADD~\cite{sladd} exhibit lower AUCs and weak generalization, confirming their limitations under complex cross-domain settings. A more detailed evaluation including additional metrics is reported in the supplementary materials.

\begin{figure*}
    \centering
    \vspace{-6mm}
    \includegraphics[width=0.6\linewidth]{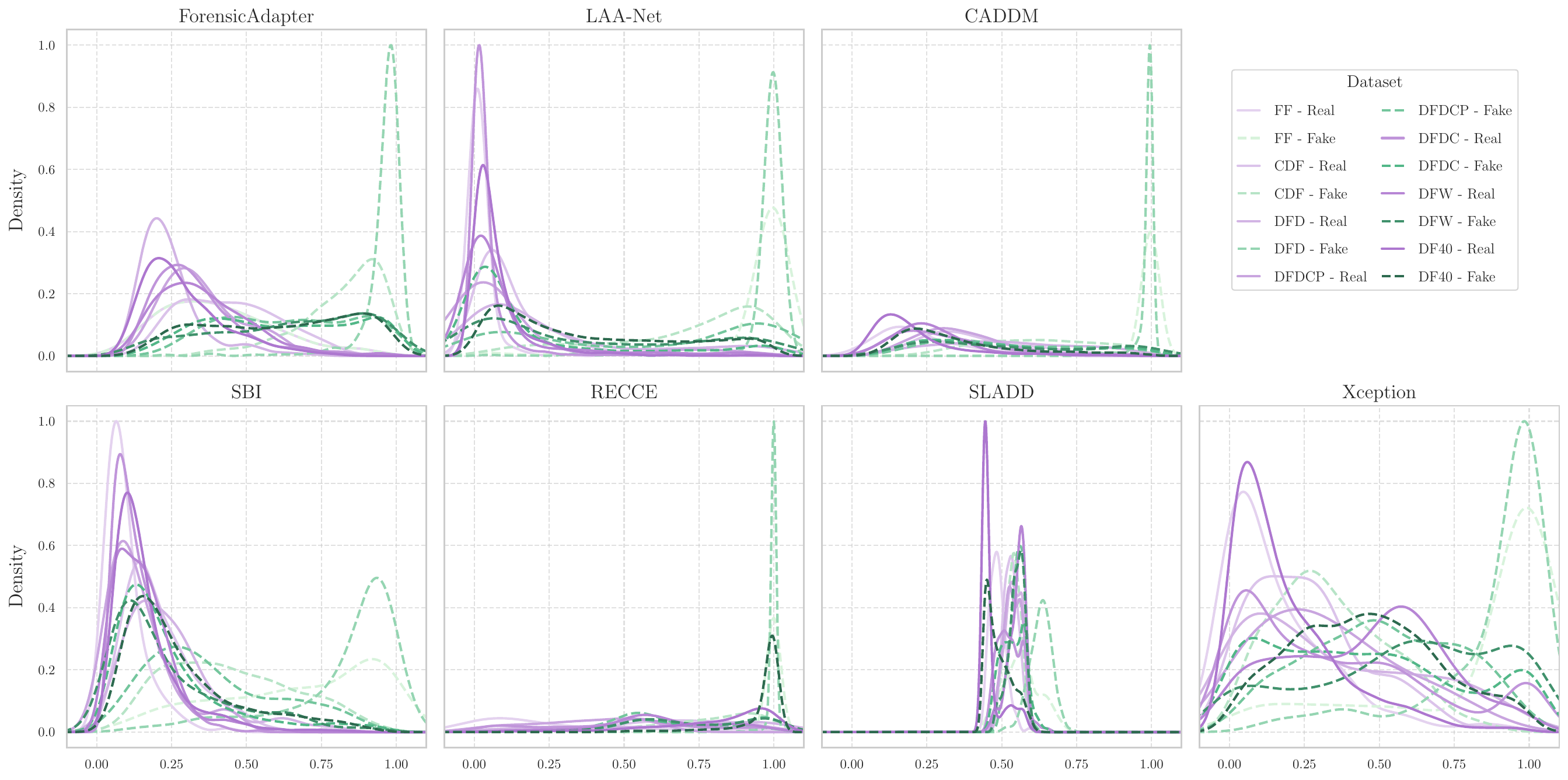}
    \vspace{-2mm}
    \caption{Class-conditional score distributions of the evaluated deepfake detectors across seven datasets~\cite{ff++, celeb_df, dfd, wdf, dfdcp, dfdc, DF40}. 
    }
    \label{fig:full_pred_patterns}
    \vspace{-6mm}
\end{figure*}

\vspace{1mm}
\noindent\textbf{Cross-AUC versus Average AUC.}
Table~\ref{table:AUC_results} compares the conventional average AUC across individual datasets (\(\text{AUC}_a\)) with the AUC measured on the combined test set (\(\text{AUC}_c\)).
Across all models, \(\text{AUC}_a\) is consistently higher than \(\text{AUC}_c\), with differences $\Delta_a$ ranging from $1.77$pp (LAA-Net) to over $12.30$pp (SLADD).
This consistent drop highlights that average AUC often overstates generalization, as it reflects performance under domain-specific conditions.
However, models in deployment encounter inputs from mixed or unknown distributions, more closely represented by the combined test set.
The lower \(\text{AUC}_c\) values expose how model predictions degrade under such conditions, revealing weaknesses that per-domain averages mask.
These results indicate that relying solely on average AUC can be misleading, highlighting the need for more realistic evaluation of generalization. 

\begin{table}
\centering
\caption{Comparison of AUC$_c$ (combined AUC) with AUC$_a$ (average per-dataset AUC), H-score~\cite{h-score}, and our Cross-AUC. $\Delta$ values are absolute deviations from AUC$_c$ with respect to other metrics (lower is better).
\boldgreen{Green} highlights the closest performance to AUC$_c$.
}
\vspace{-3mm}
\resizebox{0.66\linewidth}{!}{%
\begin{tabular}{l|c|ccc|cccHHHHHHHHHH}
\toprule
  & $\text{AUC}_{c}$ & $\text{AUC}_{a}$ & Cross-AUC & H-score & $\Delta_{a}$ & $\Delta_\text{Cross-AUC}$ & $\Delta_{\text{H-score}}$
  & $AUC_{c}$ & $AUC_{a}$ & $AUC_{p}$ & $\Delta_{a}$ & $\Delta_{p}$ & $AUC_{c}$ & $AUC_{a}$ & $AUC_{p}$ & $\Delta_{a}$ & $\Delta_{p}$ \\
\midrule
\midrule
Xception & 68.95 & 74.68 & \boldgreen{68.52} & 72.38 & 5.73 & \boldgreen{0.42} & 2.42 & 68.15 & 74.77 & \boldgreen{68.74} & 6.62 & \boldgreen{0.59} & 66.44 & 76.31 & \boldgreen{69.93} & 9.87 & \boldgreen{3.49}\\

SLADD & 57.18 & 69.48 & \boldgreen{58.37} & 65.88 & 12.30 & \boldgreen{1.19} & 8.70 & 60.36 & 70.35 & \boldgreen{59.47} & 9.99 & \boldgreen{0.88} & 58.7 & 70.82 & \boldgreen{60.93} & 12.12 & \boldgreen{2.23} \\

RECCE & 76.91 & 81.13 & 73.26 & \boldgreen{79.22} & 4.22 & 3.64 & \boldgreen{2.31} &  72.83 & 81.48 & \boldgreen{73.55} & 8.65 & \boldgreen{0.72} & 71.63 & 84.93 & \boldgreen{75.79} & 13.3 & \boldgreen{4.16} \\

SBI & 74.43 & 81.3 & \boldgreen{74.20} & 79.10 & 6.87 & \boldgreen{0.22} & 4.67 & 72.05 & 82 & \boldgreen{74.52} & 9.95 & \boldgreen{2.47} & 70.75 & 83.09 & \boldgreen{73.40} & 12.34 & \boldgreen{2.65} \\

CADDM & 71.53 & 81.91 & \boldgreen{75.31} & 80.43 & 10.38 & \boldgreen{3.78} & 8.90 & 73.09 & 82.36 & \boldgreen{76.68} & 9.27 & \boldgreen{3.59} & 72.46 & 83.4 & \boldgreen{77.26} & 10.94 & \boldgreen{4.80} \\

LAA-Net & 86.51 & 88.28 & \boldgreen{86.47} & 87.25 & 1.77 & \boldgreen{0.04} & 0.74 & 84.92 & 87.56 & \boldgreen{85.68} & 2.64 & \boldgreen{0.76} & 80.53 & 85.47 & \boldgreen{84.12} & 4.94 & \boldgreen{3.59} \\

ForensicsAdapter & 85.96 & 90.18 & \boldgreen{83.59} & 89.66 & 4.22 & \boldgreen{2.36} & 3.70 & 85.71 & 90.72 & \boldgreen{83.49} & 5.01 & \boldgreen{2.21} & 85.1 & 92.49 & \boldgreen{83.75} & 7.39 & \boldgreen{1.34} \\
\bottomrule
\end{tabular}
}
\label{table:AUC_results}
\vspace{-6mm}
\end{table}

We also analyze the gap between the predicted Cross-AUC and the combined AUC (\(\text{AUC}_c\)), as shown in Table~\ref{table:AUC_results}.
Unlike the conventional average AUC (\(\text{AUC}_a\)), which consistently overestimates performance under distribution shift, Cross-AUC is explicitly designed to estimate generalization using per-dataset AUCs and score-level polarization reported in Table~\ref{tabl:trimmed_results}.
Across all models, Cross-AUC aligns more closely with \(\text{AUC}_c\) than \(\text{AUC}_a\), achieving the smallest difference (\(\Delta_\text{Cross-AUC}\)) in all cases. In summary, Cross-AUC and average AUC imply a close yet different performance ranking of deepfake detection methods.  For instance, Table~\ref{tabl:trimmed_results} shows a rank reversal between ForensicsAdapter and LAA-Net. Although ForensicsAdapter attains the highest average AUC. LAA-Net achieves a higher combined AUC and Cross-AUC. Such cases highlight that the average AUC is not sufficient for comparing the performance of deepfake detection methods in terms of generalization.
An alternative radar-based view of the comparison is provided in the supplementary materials.

\vspace{1mm}
\noindent\textbf{Polarization Robustness under Domain Shift.} We compare the polarization scores of each model by looking at their average values across individual datasets and their value on the combined dataset, as shown in Table~\ref{tabl:trimmed_results}.
In general, most models show a lower polarization score on the combined set compared to the average across separate datasets. Specifically, LAA-Net has the highest average polarization (\eg, average $W_Q$: $0.486$) and performs well on several datasets such as CDF, DFW, DFDCP. However, it also shows the largest drop in the combined setting, with its score decreasing to $0.318$. 
On the other hand, ForensicAdapter shows a smaller decrease ($W_Q$: 0.402 to 0.336), indicating more stable score alignment across domains. SBI, CADDM, and Xception achieve strong polarity scores on specific datasets (\eg, FF++, DFD), but their combined scores are notably lower. Finally, SLADD yields low polarity in both settings ($W_Q$: 0.032 to 0.014), indicating poor class separability regardless of the domain. These observations are further supported by Figure~\ref{fig:full_pred_patterns}.

Overall, most models show a drop in polarization when moving from individual datasets to the combined set. This gap confirms that the performance of current detectors is affected by domain shifts, highlighting the importance of adopting a polarization-aware evaluation for assessing the generalization capabilities of deepfake detectors.

\vspace{1mm}
\noindent\textbf{Cross-AUC versus Generalization Metrics.} To further demonstrate the effectiveness of Cross-AUC, we additionally compare it with the H-score~\cite{h-score} metric in Table~\ref{table:AUC_results}. H-score was originally designed for universal domain adaptation and represents the harmonic mean of accuracy on common and unknown classes. To align with our setting, we replace accuracy with per-dataset AUC. The obtained results demonstrate that Cross-AUC is more clearly correlated with the combined AUC.

\begin{wraptable}{r}{0.55\linewidth}
\centering
\vspace{-1.1cm}
\caption{Estimated Cross-AUC with different $\Psi$ functions, \ie, harmonic, arithmetic, and geometric means. The difference ($\Delta$) between the AUC$_c$ and the resulting Cross-AUC is reported for each function.}
\resizebox{\linewidth}{!}{%
\begin{tabular}{l|c|cc|cc|cc}
  \toprule
\multirow{2}{*}{Model} & \multirow{2}{*}{$\text{AUC}_c$} & \multicolumn{6}{c}{Cross-AUC} \\
\cline{3-8}
& & \rule{0pt}{2.3ex}$\Psi_{\text{Harmonic}}$ & $\Delta$ & $\Psi_{\text{Arithmetic}}$ & $\Delta$ & $\Psi_{\text{Geometric}}$ & $\Delta$ \\
\midrule
\midrule
    Xception & 68.95 &        \boldgreen{68.52} & \boldgreen{0.42}  &        69.93 &  0.98 &      67.29 & 1.66\\
    SLADD & 57.18 &       \boldgreen{58.38} & \boldgreen{1.19} &         60.80 & 3.61 &       59.72 & 2.53 \\
    RECCE & 76.91 &       73.27 & 3.64 &        \boldgreen{77.40} & \boldgreen{0.49} &       73.98 & 2.92\\
    SBI & 74.43 &       \boldgreen{74.21} &   \boldgreen{0.22}&       77.45 &  3.02 &       73.18 & 1.24\\
    CADDM & 71.53 &       \boldgreen{75.32} &    \boldgreen{3.78} &     80.61 & 9.07 &       77.54 & 6.00 \\
    LAA-Net & 86.51 &       \boldgreen{86.47} &   \boldgreen{0.04} &      82.52 & 3.98 &         85.37 & 1.13 \\
    ForensicsAdapter & 85.96 &       \boldgreen{83.60} &   \boldgreen{2.36} &      82.02 &  3.94 &      82.94 & 3.02 \\
\bottomrule
\end{tabular}
}
\vspace{-7mm}
\label{tab:mean_aucs}
\end{wraptable}

\vspace{1mm}
\noindent\textbf{Effect of $\Psi$.}
Table~\ref{tab:mean_aucs} compares three averaging functions $\Psi$, including harmonic, arithmetic, and geometric, used compute Cross-AUC in Eq.~(\ref{eq:cross_auc}).
We evaluate their effectiveness by measuring the absolute difference from the combined AUC (\(\text{AUC}_c\)). We evaluate their effectiveness by measuring the absolute difference between Cross-AUC and combined AUC (\(\text{AUC}_c\)).
The results show that the harmonic mean consistently produces the closest results, achieving the lowest difference (\(\Delta\)) in $6$ out of $7$ models. For that reason, we adopt the harmonic mean in the rest of our experiments.





\begin{table}[]
\centering
\vspace{-6mm}
\caption{Effects of the balancing factor $\lambda$.}
\vspace{-3mm}
\resizebox{0.73\linewidth}{!}{
\begin{tabular}{c|c|cccccccccc}
    \toprule
         & \multirow{2}{*}{$\text{AUC}_c$} & \multicolumn{10}{c}{Cross-AUC} \\
         \cline{3-12}
         &  & $\lambda$=0.1 & 0.2 & 0.3 & 0.4 & 0.5 & 0.6 & 0.7 & 0.8 & 0.9 & 1.0 \\
         \midrule
         \midrule
         Xception & 68.95 & 71.61 & 70.83 & 70.06 & \boldgreen{69.29} & 68.52 & 67.75 & 66.98 & 66.20 & 65.43 & 64.66 \\
         SLADD & 57.18 & 64.37 & 62.87 & 61.37 & 59.87 & 58.37 & \boldgreen{56.87} & 55.37 & 53.87 & 52.37 & 50.86 \\
         RECCE & 76.91 & 78.03 & \boldgreen{76.83} & 75.64 & 74.45 & 73.26 & 72.07 & 70.88 & 69.69 & 68.50 & 67.30 \\
         SBI & 74.43 & 78.12 & 77.14 & 76.16 & 75.18 & \boldgreen{74.20} & 73.22 & 72.24 & 71.26 & 70.28 & 69.30 \\
         CADDM & 71.53 & 79.40 & 78.38 & 77.36 & 76.34 & 75.31 & 74.29 & 73.27 & 72.25 & \boldgreen{71.22} & 70.20 \\
         LAA-Net & 86.51 & 87.09 & 86.94 & 86.78 & 86.62 & \boldgreen{86.47} & 86.30 & 86.15 & 85.99 & 85.83 & 85.67 \\
         ForensicsAdapter & 85.96 & 88.45 & 87.23 & \boldgreen{86.02} & 84.81 & 83.59 & 82.38 & 81.16 & 79.95 & 78.74 & 77.52 \\
         \bottomrule
\end{tabular}
}
\label{tab:different_lambda}
\vspace{-5mm}
\end{table}

\noindent\textbf{Effect of the Balancing Factor $\lambda$.}
We also analyze the sensitivity of Cross-AUC to the balancing factor $\lambda$ defined in Eq.~(\ref{eq:cross_auc}). As shown in Table~\ref{tab:different_lambda}, setting $\lambda$ to very small values deviates Cross-AUC from the combined AUC, making it closer to a simple average AUC. On the other hand, the use of excessively large values for $\lambda$ overweights polarization. A mid-range choice ($\lambda$ $\approx$ $0.5$) provides Cross-AUC values that closer to AUC$_c$.

\begin{table}
\centering
\vspace{-5mm}
\caption{Estimated Cross-AUC with different polarization estimators, \ie, Wasserstein Distance (WD), KL Divergence (KL), and Jensen-Shannon (JS) Distance. }
\vspace{-3mm}
\resizebox{0.7\linewidth}{!}{%
\begin{tabular}{c|ccccccc|c}
\toprule
  & Xception & SLADD & RECCE & SBI & CADDM & LAA-Net & ForensicsAdapter & Avg. \\
\midrule
\midrule
AUC$_c$ & 68.95 & 57.18 & 76.91 & 74.43 & 71.53 & 86.51 & 85.96 & 74.49 \\
\midrule
Cross-AUC$_{\text{WD}}$ & \boldgreen{68.52} & \boldgreen{58.37} & 73.26 & \boldgreen{74.20} & \boldgreen{75.31} & 86.47 & 83.59 & \boldgreen{74.24} \\
$\text{Cross-AUC}_{\text{KL}}$ & 68.29 & 60.07 & \boldgreen{74.22} & 73.88 & 76.80 & \boldgreen{86.54} & \boldgreen{87.05} & 75.26 \\
$\text{Cross-AUC}_{\text{JS}}$ & 54.61 & 65.73 & 60.82 & 66.05 & 62.22 & 67.48 & 63.45 & 62.90 \\
\bottomrule
\end{tabular}
}
\label{table:different_estimators}
\vspace{-5mm}
\end{table}

\vspace{1mm}
\noindent\textbf{Cross-AUC with Different Polarization Estimator $\Pi$.} 
We examine the effect of replacing the WD in Cross-AUC with alternative  measures, including KL Divergence (KL) and Jensen-Shannon Distance (JS). As shown in Table~\ref{table:different_estimators}. Both WD and KL-based Cross-AUC remain closely aligned with the combined AUC, whereas the JS variant yields less consistent estimates. This suggests that while Cross-AUC is generally robust to the choice of estimator, WD offers the most stable results under distribution shift.

\vspace{1mm}

\begin{wraptable}{r}{0.5\linewidth}
\centering
\vspace{-1.1cm}
\caption{An out-of-distribution scenario with MagicBrush~\cite{magicbrush}. 
}
\resizebox{\linewidth}{!}{
\begin{tabular}{c| c| cccc}
    \toprule
         & & MagicBrush & AUC$_c$ & AUC$_a$ & Cross-AUC \\
    \midrule
    \midrule
    \multirow{2}{*}{Xception} & AUC & 44.90 & 66.70 & 70.96 & \boldgreen{65.54} \\
             & $W_Q$ & 0.029 & 0.170 & 0.246 & \\
    \midrule
    \multirow{2}{*}{SLADD} & AUC & 38.25 & 56.98 & 65.58 & \boldgreen{58.44} \\
          & $W_Q$ & 0.007 & 0.013 & 0.029 & \\
    \midrule
    \multirow{2}{*}{RECCE} & AUC & 53.70 & 72.23 & 77.70 & \boldgreen{73.20} \\
          & $W_Q$ & 0.001 & 0.196 & 0.239 & \\
    \midrule
    \multirow{2}{*}{SBI} & AUC & 86.71 & 72.79 & 81.98 & \boldgreen{78.65} \\
        & $W_Q$ & 0.209 & 0.187 & 0.269 & \\
    \midrule
    \multirow{2}{*}{CADDM} & AUC & 53.24 & 69.60 & 78.33 & \boldgreen{74.08} \\
          & $W_Q$ & 0.012 & 0.172 & 0.272 & \\
    \midrule
    \multirow{2}{*}{LAA-Net} & AUC & 55.73 & 85.26 & \boldgreen{84.21} & 80.03 \\
            & $W_Q$ & 0.018 & 0.309 & 0.428 & \\
    \midrule
    \multirow{2}{*}{ForensicsAdapter} & AUC & 59.42 & 83.22 & 86.34 & \boldgreen{83.01} \\
                     & $W_Q$ & 0.061 & 0.298 & 0.359 & \\
    \bottomrule
\end{tabular}
}
\label{tab:ood}
\vspace{-6mm}
\end{wraptable}

\noindent\textbf{OOD Evaluation with MagicBrush.}
We further evaluate the relevance of Cross-AUC beyond facial deepfake detection. Specifically, we consider an Out-Of-Distribution (OOD) detection task using the MagicBrush~\cite{magicbrush} dataset, which contains non-facial manipulations that are unseen in other benchmarks. We report the obtained results in Table~\ref{tab:ood}. It shows that the gap between the average AUC and the combined AUC is even more pronounced, while Cross-AUC remains consistently close to the combined AUC. This highlights that Cross-AUC provides a reliable measure for both facial and non-facial content.

\vspace{1mm}
\noindent\textbf{Limitations.}
While Cross-AUC is simple, effective, and practical, we acknowledge the lack of theoretically grounded justifications. Despite this issue, we believe that this work remains valuable for the community. In fact,  it highlights a persistent problem in deepfake detection that is mostly ignored in the literature, which hinders the transition toward truly deployable solutions.  Moreover, it suggests, as a potential solution, the exploitation of polarization. Another limitation of this work lies in its exclusive focus on the deepfake detection use case. However, the observed lack of robustness of AUC under domain shift is, in principle, not specific to this application. It represents a broader issue that can affect a wide range of binary classification tasks. Studying Cross-AUC for other binary classification tasks will be, therefore, investigated in the future.





\vspace{-2mm}
\section{Conclusion}
\label{sec:cc}
\vspace{-2mm}
This work addresses the need to assess the generalization capabilities of deepfake detectors more accurately, as current evaluation practices overlook the complexity of real-world scenarios.
We empirically demonstrate that computing AUC scores separately across datasets obscures the true capabilities of detection models, which are expected to perform on real-world, mixed-domain data.
To address this, we characterize the polarization of deepfake detector models by quantifying the degree of separation between real and fake score distributions.
We introduce Cross-dataset AUC (Cross-AUC), a metric that extends the AUC scores by explicitly incorporating these score polarizations.
This approach better captures the robustness of the models under domain shifts, which is expected in realistic evaluations.
Our experiments indicate that most state-of-the-art detectors fail to generalize as much as they claim, and that Cross-AUC is a practical metric for evaluating their generalization performance, thereby highlighting the relevance of emphasizing polarity.

\bibliographystyle{splncs04}
\bibliography{main}

\newpage
\section{Appendix}
\label{sec:appendix}

Our technical appendix is organized as follows.
Section~\ref{subsec:additional_results} presents additional results and extended analyses that complement Section~\textcolor{RubineRed}{4} of the main paper.
Section~\ref{subsec:dataset_info} details the dataset information and selection criteria.
Section~\ref{subsec:detectors} describes the seven deepfake detectors used in our experiments.
Finally, Section~\ref{subsec:metrics} summarizes additional evaluation metrics beyond AUC and polarization.



{\tiny        
\begin{longtable}{ll|c| *{8}{c}}
\caption{Performance of the selected models on all datasets. 
           \textbf{$\tau$}: Optimal threshold; 
           $\phi_{\tau}$: variance of the top–\(K\) optimal thresholds in each dataset;
           $W_\text{KDE}$, $W_\text{Q}$, $W_\text{GMM}$, $W_\text{BD}$: model polarities using WD as $\Pi$ over different density estimation ($\Theta$) methods, including non-parametric (Kernel Density Estimation (KDE), Quantile (Q)), and parametric ones assuming two types of distributions, \ie, GMM and BD.
           \boldgray{Gray} represents the dataset the model was trained on. \boldgreen{Green} highlights the best performance and polarities.}
\label{tabl:full_res} \\
\toprule
 & & \textcolor{gray}{FF++} & CDF & DFD & DFW & DFDCP & DFDC & DF40 & Average & Combined \\
 \midrule
 \midrule
 \endfirsthead

 \toprule
 & & \textcolor{gray}{FF++} & CDF & DFD & DFW & DFDCP & DFDC & DF40 & Average & Combined \\
 \midrule
 \midrule
 \endhead

 \bottomrule
 \endfoot

 \bottomrule
 \endlastfoot
 
 \multirow{13}{*}{Xception~\cite{xception}} & ACC (\%) & \textcolor{gray}{85.32} & 43.19 & 82.59 & 58.06 & 53.44 & 55.92 & 57.56 & 62.29 & 58.53
\\ & BACC (\%) & \textcolor{gray}{87.31} & 53.55 & 79.79 & 57.95 & 63.7 & 55.84 & 67.17 & 66.47 & 61.6
\\ & AUC (\%) & \textcolor{gray}{93.6} & 62.1 & 92.24 & 63.71 & 70.56 & 58.98 & 81.59 & 74.68 & 68.95 ($\textcolor{red}{\downarrow 5.73}$)
\\ & P (\%) & \textcolor{gray}{96.03} & 74.48 & 83.24 & 57.8 & 94.54 & 57.71 & 41.95 & 72.25 & 75.2
\\ & R (\%) & \textcolor{gray}{81.34} & 21.72 & 90.58 & 64.62 & 50.48 & 42.39 & 92.49 & 63.37 & 47.26 
\\ & S (\%) & \textcolor{gray}{93.28} & 85.38 & 69 & 51.28 & 76.92 & 69.29 & 92.4 & 76.79 & 75.94 
\\ & F1 (\%) & \textcolor{gray}{88.08} & 33.64 & 86.76 & 61.02 & 65.82 & 48.88 & 57.72 & 63.13 & 58.04
\\ & EER (\%) & \textcolor{gray}{13.43} & 42.69 & 15.5 & 40.51 & 33.33 & 43.58 & 25.6 & 30.66 & 35.68 (\textcolor{red}{$\uparrow$ 5.02}) 
\\ & $\tau$ & \textcolor{gray}{0.4} & 0.24 & 0.7 & 0.57 & 0.4 & 0.3 & 0.24 & 0.4 & 0.31 
\\ & $\phi_{\tau}$ & \textcolor{gray}{0.0004} & 0.0004& 0.0002& 0.0007& 0.0010& 0.0001 & 0.0001 & 0.0004& 0.0001
\\ & $W_{\text{KDE}}$ & \textcolor{gray}{0.524} & 0.064 & 0.430 & 0.110 & 0.146 & 0.078 & 0.239 & 0.227 & 0.168 
\\ & $W_{\text{BD}}$ & \textcolor{gray}{0.549} & 0.068 & 0.446 & 0.122 & 0.164 & 0.060 & 0.232 & 0.234 & 0.152
\\ & $W_{\text{Q}}$ & \textcolor{gray}{0.661} & 0.077 & 0.524 & 0.139 & 0.183 & 0.091 & 0.264 & 0.277 & 0.190
\\ & $W_{\text{GMM}}$ & \textcolor{gray}{0.616} & 0.069 & 0.496 & 0.149 & 0.118 & 0.081 & 0.249 & 0.254 & 0.180
\\
\midrule
\multirow{13}{*}{SLADD~\cite{sladd}} & ACC (\%) & \textcolor{gray}{81.84} & 64.49 & 65.92 & 51.15 & \boldgreen{87.06} & 53.75 & 50.9 & 65.01 & 55.77 
\\ & BACC (\%) & \textcolor{gray}{80.03} & 50.95 & 54 & 50.45 & 53.51 & 53.9 & 59.55 & 57.48 & 53.38 
\\ & AUC (\%) & \textcolor{gray}{88.31} & 56.65 & 97.22 & 48.63 & 65.98 & 57.37 & 72.25 & 69.48 & 57.18 ($\textcolor{red}{\downarrow 12.3}$)
\\ & P (\%) & \textcolor{gray}{87.07} & 66.73 & 64.8 & 51.05 & 89.52 & 52.3 & 82.25 & 70.53 & 63.29
\\ & R (\%) & \textcolor{gray}{85.44} & 92.55 & \boldgreen{100} & \boldgreen{93.65} & \boldgreen{96.76} & 79.14 & 36.86 & 83.48 & 64.56 
\\ & S (\%) & \textcolor{gray}{74.62} & 9.35 & 8 & 7.25 & 10.25 & 28.65 & 82.24 & 31.48 & 42.2 
\\ & F1 (\%) & \textcolor{gray}{86.25} & 77.55 & 78.7 & 66.08 & \boldgreen{93} & 62.98 & 50.91 & 73.63 & 63.92 
\\ & EER (\%) & \textcolor{gray}{18.65} & 44.44 & 7.5 & 52.22 & 41.02 & 45.21 & 29.59 & 34.09 & 46.4 (\textcolor{red}{$\uparrow$ 12.31})
\\ & $\tau$ & \textcolor{gray}{0.51} & 0.53 & 0.57 & 0.55 & 0.53 & 0.53 & 0.45 & 0.52 & 0.52 
\\ & $\phi_{\tau}$ & \textcolor{gray}{0.0001} & 0.0001& 0.0001 & 0.0001& 0.0001& 0.0001 & 0.0001 & 0.0001& 0.0001
\\ & $W_{\text{KDE}}$ & \textcolor{gray}{0.060} & 0.001 & 0.098 & 0.004 & 0.022 & 0.012 & 0.003 & 0.028 & 0.013 
\\ & $W_{\text{BD}}$ & \textcolor{gray}{0.078} & 0.003 & 0.079 & 0.001 & 0.007 & 0.008 & 0.024 & 0.029 & 0.013 
\\ & $W_{\text{Q}}$ & \textcolor{gray}{0.079} & 0.005 & 0.090 & 0.002 & 0.015 & 0.010 & 0.026 & 0.032 & 0.014
\\ & $W_{\text{GMM}}$ & \textcolor{gray}{0.079} & 0.005 & 0.089 & 0.004 & 0.015 & 0.010 & 0.027 & 0.033 & 0.014
\\ 
\midrule
\multirow{13}{*}{RECCE~\cite{ete_recons}} & ACC (\%) & \textcolor{gray}{92.28} & 70.01 & 75.18 & 54.6 & 75.28 & 63.4 & 71.21 & 71.7 & 68.12 
\\ & BACC (\%) & \textcolor{gray}{88.8} & 58.57 & 66.5 & 54.05 & 70.4 & 63.52 & 53.89 & 65.1 & 61.4 
\\ & AUC (\%) & \boldgray{99.56} & 69.02 & 97.82 & 65.19 & 82.24 & 71.19 & 82.94 & 81.13 & 76.91 ($\textcolor{red}{\downarrow 4.22}$) 
\\ & P (\%) & \textcolor{gray}{90.16} & 70.62 & 71.72 & 53.2 & 94.42 & 59.25 & 70.78 & 72.87 & 67.17 
\\ & R (\%) & \boldgray{99.25} & 93.75 & \boldgreen{100} & 88.43 & 76.69 & \boldgreen{84.46} & \boldgreen{99.31} & \boldgreen{91.69} & \boldgreen{92.82} 
\\ & S (\%) & \textcolor{gray}{78.35} & 23.39 & 33 & 19.67 & 64.1 & 42.58 & 8.48 & 38.51 & 29.98 
\\ & F1 (\%) & \textcolor{gray}{94.49} & 80.56 & 83.53 & 66.43 & 84.64 & 69.64 & 82.65 & 80.27 & \boldgreen{77.94} 
\\ & EER (\%) & \boldgray{1.49} & 36.25 & 8 & 39.81 & 28.2 & 34.43 & 24.48 & 24.66 & 31.85 (\textcolor{red}{$\uparrow$ 7.19}) 
\\ & $\tau$ & \textcolor{gray}{0.98} & 0.74 & 0.99 & 0.65 & 0.53 & 0.58 & 0.97 & 0.77 & 0.86 
\\ & $\phi_{\tau}$ & \textcolor{gray}{0.00001} & 0.0002& 0.0001& 0.0001& 0.0001& 0.0001 & 0.0001& 0.0001& 0.0001
\\ & $W_{\text{KDE}}$ & \textcolor{gray}{0.631} & 0.126 & 0.352 & 0.096 & 0.212 & 0.181 & 0.153 & 0.250 & 0.221 
\\ & $W_{\text{BD}}$ & \textcolor{gray}{0.569} & 0.124 & 0.280 & 0.116 & 0.240 & 0.182 & 0.117 & 0.233 & 0.201 
\\ & $W_{\text{Q}}$ & \textcolor{gray}{0.716} & 0.136 & 0.355 & 0.115 & 0.243 & 0.203 & 0.144 & 0.273 & 0.232
\\ & $W_{\text{GMM}}$ & \textcolor{gray}{0.692} & 0.134 & 0.349 & 0.111 & 0.227 & 0.203 & 0.142 & 0.265 & 0.228 
\\ 
\midrule
\multirow{13}{*}{SBI~\cite{sbi}} & ACC (\%) & \textcolor{gray}{80.59} & 58.57 & 90.92 & 55.06 & 35.91 & 55.12 & 40.14 & 59.47 & 51.7 
\\ & BACC (\%) & \textcolor{gray}{85.26} & 68.17 & 91.35 & 55.76 & 63.91 & 54.86 & 55.98 & 67.89 & 59.95 
\\ & AUC (\%) & \textcolor{gray}{98.23} & 84.65 & 96.78 & 60.83 & 86.83 & 69.77 & 72.05 & 81.3 & 74.43 ($\textcolor{red}{\downarrow 6.87}$) 
\\ & P (\%) & \boldgray{99.47} & 97.01 & 95.61 & \boldgreen{91.8} & \boldgreen{100} & \boldgreen{97.9} & 92.86 & \boldgreen{96.37} & 95.84 
\\ & R (\%) & \textcolor{gray}{71.26} & 38.69 & 89.7 & 12.69 & 27.83 & 9.94 & 14.44 & 37.79 & 21.34 
\\ & S (\%) & \boldgray{99.25} & \boldgreen{97.66} & 93 & \boldgreen{98.82} & \boldgreen{100} & \boldgreen{99.78} & 97.52 & \boldgreen{98} & \boldgreen{98.57}
\\ & F1 (\%) & \textcolor{gray}{83.04} & 55.31 & 92.56 & 22.31 & 43.54 & 18.05 & 24.99 & 48.54 & 34.9 
\\ & EER (\%) & \textcolor{gray}{6.71} & 24.56 & 9 & 46.83 & 20.51 & 36.11 & 33.36 & 25.29 & 32.57 (\textcolor{red}{$\uparrow$ 7.28})
\\ & $\tau$ & \textcolor{gray}{0.21} & 0.28 & 0.46 & 0.16 & 0.19 & 0.14 & 0.17 & 0.23 & 0.17 
\\ & $\phi_{\tau}$ & \textcolor{gray}{0.0006} & 0.0001& 0.0008& 0.0001& 0.0002& 0.0001 & 0.0001& 0.0002& 0.0001
\\ & $W_{\text{KDE}}$ & \textcolor{gray}{0.503} & 0.215 & 0.516 & 0.078 & 0.220 & 0.097 & 0.103 & 0.248 & 0.152 
\\ & $W_{\text{BD}}$ & \textcolor{gray}{0.513} & 0.219 & 0.500 & 0.089 & 0.224 & 0.116 & 0.112 & 0.253 & 0.182 
\\ & $W_{\text{Q}}$ & \textcolor{gray}{0.576} & 0.238 & 0.577 & 0.085 & 0.244 & 0.108 & 0.116 & 0.278 & 0.172 
\\ & $W_{\text{GMM}}$ & \textcolor{gray}{0.567} & 0.236 & 0.564 & 0.084 & 0.245 & 0.105 & 0.115 & 0.274 & 0.171
\\ 
\midrule
\multirow{13}{*}{CADDM~\cite{caddm}} & ACC (\%) & \textcolor{gray}{95.27} & 72.19 & 80.74 & 69.01 & 52.01 & 63.72 & 43.7 & 68.09 & 58.55 
\\ & BACC (\%) & \textcolor{gray}{93.47} & 72.56 & 74 & 69.35 & 68.5 & 63.6 & 57.41 & 71.27 & 63.32 
\\ & AUC (\%) & \textcolor{gray}{99.26} & 80.7 & 99.52 & 76.31 & 71 & 70.33 & 76.28 & 81.91 & 71.53 ($\textcolor{red}{\downarrow 10.38}$) 
\\ & P (\%) & \textcolor{gray}{94.31} & 84.21 & 76.58 & 84.13 & 97.33 & 72.92 & 87.83 & 85.33 & 81.51 
\\ & R (\%) & \textcolor{gray}{98.88} & 71.43 & \boldgreen{100} & 48.07 & 47.25 & 42.97 & 21.46 & 61.43 & 41.01 
\\ & S (\%) & \textcolor{gray}{88.06} & 73.68 & 48 & 90.63 & 89.75 & 84.22 & 93.36 & 81.1 & 85.64 
\\ & F1 (\%) & \textcolor{gray}{96.54} & 77.3 & 86.73 & 61.18 & 63.61 & 54.08 & 34.49 & 67.7 & 54.56 
\\ & EER (\%) & \textcolor{gray}{2.24} & 27.49 & 3.5 & 30.91 & 38.46 & 36.06 & 30.48 & 24.16 & 35.97 (\textcolor{red}{$\uparrow$ 11.81})
\\ & $\tau$ & \textcolor{gray}{0.97} & 0.45 & 0.93 & 0.37 & 0.44 & 0.37 & 0.22 & 0.53 & 0.31 
\\ & $\phi_{\tau}$ & \textcolor{gray}{0.0253} & 0.0001& 0.0001& 0.0001& 0.0011& 0.0001 & 0.0001& 0.0038& 0.0001
\\ & $W_{\text{KDE}}$ & \textcolor{gray}{0.661} & 0.203 & 0.436 & 0.210 & 0.156 & 0.155 & 0.137 & 0.280 & 0.176 
\\ & $W_{\text{BD}}$ & \textcolor{gray}{0.607} & 0.203 & 0.386 & 0.238 & 0.182 & 0.165 & 0.146 & 0.275 & 0.214 
\\ & $W_{\text{Q}}$ & \textcolor{gray}{0.726} & 0.226 & 0.466 & 0.246 & 0.179 & 0.174 & 0.151 & 0.310 & 0.198 
\\ & $W_{\text{GMM}}$ & \textcolor{gray}{0.714} & 0.224 & 0.459 & 0.241 & 0.174 & 0.173 & 0.150 & 0.305 & 0.197 
\\ 
\midrule
\multirow{13}{*}{LAA-Net~\cite{laa_net}} & ACC (\%) & \boldgray{96.67} & 84.72 & 92.5 & 64.52 & 59.74 & 57.68 & 37.79 & 70.51 & 45.35 
\\ & BACC (\%) & \boldgray{97.14} & \boldgreen{87.71} & 88.67 & 65.01 & 72.36 & 57.37 & 65.99 & 76.32 & 66.15 
\\ & AUC (\%) & \textcolor{gray}{99.45} & \boldgreen{95.45} & 98.47 & 79.52 & 86.48 & 72.6 & 86.04 & 88.28 & \boldgreen{86.51} ($\textcolor{red}{\downarrow 1.77}$)
\\ & P (\%) & \textcolor{gray}{99.26} & \boldgreen{98.15} & 91.71 & 85.22 & 97.4 & 90.99 & \boldgreen{99.66} & 94.62 & \boldgreen{98.27} 
\\ & R (\%) & \textcolor{gray}{95.71} & 78.24 & 98.19 & 36.59 & 55.83 & 16.33 & 33.63 & 59.21 & 35.59 
\\ & S (\%) & \textcolor{gray}{98.57} & 97.18 & 79.15 & 93.43 & 88.89 & 98.41 & \boldgreen{98.35} & 93.42 & 96.71 
\\ & F1 (\%) & \boldgray{97.45} & 87.07 & 94.84 & 51.19 & 70.98 & 27.69 & 50.29 & 68.5 & 52.25 
\\ & EER (\%) & \textcolor{gray}{2.86} & \boldgreen{11.86} & 5.96 & 27.27 & \boldgreen{18.52} & 33.91 & \boldgreen{22.2} & 17.51 & \boldgreen{21.4} (\textcolor{red}{$\uparrow$ 3.89})
\\ & $\tau$ & \textcolor{gray}{0.23} & 0.3 & 0.96 & 0.06 & 0.08 & 0.03 & 0.11 & 0.25 & 0.083 
\\ & $\phi_{\tau}$ & \textcolor{gray}{0.0161} & 0.0004& 0.0001& 0.0001& 0.0005& 0.0001& 0.0001& 0.0024& 0.0001
\\ & $W_{\text{KDE}}$ & \boldgray{0.831} & \boldgreen{0.511} & 0.633 & \boldgreen{0.262} & \boldgreen{0.334} & 0.169 & 0.281 & \boldgreen{0.431} & 0.302 
\\ & $W_{\text{BD}}$ & \boldgray{0.749} & \boldgreen{0.495} & 0.547 & 0.248 & \boldgreen{0.357} & 0.189 & 0.272 & \boldgreen{0.408} & 0.296 
\\ & $W_{\text{Q}}$ & \boldgray{0.927} & \boldgreen{0.582} & 0.708 & \boldgreen{0.285} & \boldgreen{0.446} & 0.156 & 0.299 & \boldgreen{0.486} & 0.318 
\\ & $W_{\text{GMM}}$ & \boldgray{0.911} & \boldgreen{0.579} & 0.690 & \boldgreen{0.292} & \boldgreen{0.417} & 0.154 & 0.286 & \boldgreen{0.475} & 0.317 
\\
\midrule
\multirow{13}{*}{ForensicAdapter~\cite{forensicsadapter}} & ACC (\%) & \textcolor{gray}{89.3} & \boldgreen{85.21} & \boldgreen{97.96} & \boldgreen{76.61} & 70.11 & \boldgreen{75.46} & \boldgreen{71.96} & \boldgreen{80.94} & \boldgreen{76.19} 
\\ & BACC (\%) & \textcolor{gray}{84.89} & 79.94 & \boldgreen{97.56} & \boldgreen{76.72} & \boldgreen{75.33} & \boldgreen{75.39} & \boldgreen{77.12} & \boldgreen{80.99} & \boldgreen{78.15} 
\\ & AUC (\%) & \textcolor{gray}{98.39} & 94.94 & \boldgreen{99.8} & \boldgreen{80.95} & \boldgreen{87.04} & \boldgreen{83.88} & \boldgreen{86.3} & \boldgreen{90.18} & 85.96 ($\textcolor{red}{\downarrow 4.22}$)
\\ & P (\%) & \textcolor{gray}{87.38} & 83.9 & \boldgreen{97.68} & 81.32 & 96.8 & 84.2 & 93.82 & 89.3 & 89.36 
\\ & R (\%) & \textcolor{gray}{98.13} & \boldgreen{96.13} & 99.11 & 70.07 & 68.61 & 62.34 & 63.6 & 79.71 & 68.99
\\ & S (\%) & \textcolor{gray}{71.64} & 63.74 & \boldgreen{96} & 83.37 & 82.05 & 88.43 & 90.64 & 82.26 & 87.32 
\\ & F1 (\%) & \textcolor{gray}{92.44} & \boldgreen{89.6} & \boldgreen{98.39} & \boldgreen{76.27} & 80.3 & \boldgreen{71.64} & 75.8 & \boldgreen{83.49} & 77.86 
\\ & EER (\%) & \textcolor{gray}{5.22} & 12.28 & \boldgreen{1.5} & \boldgreen{25.53} & 23.08 & \boldgreen{24.03} & 22.32 & \boldgreen{16.28} & 22.39 (\textcolor{red}{$\uparrow$ 6.11})
\\ & $\tau$ & \textcolor{gray}{0.76} & 0.64 & 0.65 & 0.5 & 0.43 & 0.4 & 0.38 & 0.53 & 0.43
\\ & $\phi_{\tau}$ & \textcolor{gray}{0.0013} & 0.0003& 0.0027& 0.0001& 0.0007& 0.0001& 0.0001& 0.0007& 0.0001
\\ & $W_{\text{KDE}}$ & \textcolor{gray}{0.497} & 0.338 & \boldgreen{0.659} & 0.251 & 0.259 & \boldgreen{0.257} & \boldgreen{0.291} & 0.365 & \boldgreen{0.304}
\\ & $W_{\text{BD}}$ & \textcolor{gray}{0.460} & 0.334 & \boldgreen{0.624} & \boldgreen{0.253} & 0.268 & \boldgreen{0.266} & \boldgreen{0.294} & 0.357 & \boldgreen{0.306}
\\ & $W_{\text{Q}}$ & \textcolor{gray}{0.540} & 0.374 & \boldgreen{0.713} & 0.285 & 0.293 & \boldgreen{0.286} & \boldgreen{0.322} & 0.402 & \boldgreen{0.336}
\\ & $W_{\text{GMM}}$ & \textcolor{gray}{0.521} & 0.371 & \boldgreen{0.709} & 0.284 & 0.288 & \boldgreen{0.285} & \boldgreen{0.320} & 0.397 & \boldgreen{0.335} \\
\bottomrule
\end{longtable}
}

\subsection{Additional Results and Discussions}
\label{subsec:additional_results}

In addition to AUC, optimal threshold ($\tau$), threshold variance ($\phi_\tau$), and polarization scores are reported in Table~\textcolor{RubineRed}{2} of the main paper. In this section, for the sake of a more in-depth analysis, we provide results using additional standard classification metrics, namely, Accuracy (ACC), Balanced Accuracy (BACC), Precision (P), Recall (R), Specificity (S), F1-score (F1), and Equal Error Rate (EER). Table~\ref{tabl:full_res} presents the results of the seven considered detection models~\cite{ff++, sladd, ete_recons, sbi, caddm, laa_net, forensicsadapter}, all trained on FF++~\cite{ff++}, across six unseen datasets~\cite{celeb_df, dfd, wdf, dfdc, dfdcp, DF40} and the Combined test set. We note that we set a fixed threshold of $0.5$ to compute these classification metrics.

\vspace{2mm}
\noindent\textbf{Conventional Cross-Dataset Generalization Evaluation.}
This setup directly tests cross-dataset generalization, a key challenge in deepfake detection, where models often fail when faced with manipulations or domains not seen during training.

It can be observed that results show substantial performance variation across datasets. ForensicAdapter~\cite{forensicsadapter} consistently performs well across all test sets, maintaining strong ACC, AUC, and F1, which reflects robust generalization. RECCE~\cite{ete_recons} also performs well, especially in R, achieving high values on datasets like DFD~\cite{dfd} and DF40~\cite{DF40}. However, its AUC drops below $70\%$ on CDF~\cite{celeb_df} and DFW~\cite{wdf}, indicating that its generalization is not uniformly strong across all domains. LAA-Net~\cite{laa_net} achieves very high AUC on some datasets, such as CDF ($95.45\%$) and DFD ($98.47\%$), but its performance decreases on others like DFDC~\cite{dfdc} ($72.60\%$) and DF40 ($86.04\%$), showing it is more sensitive to certain shifts. CADDM~\cite{caddm} performs moderately overall, with solid AUC and accuracy on several datasets, but lower R and higher EER on DFDC and DF40 indicate weaker transfer to more challenging domains.

In contrast, SLADD~\cite{sladd} and Xception~\cite{ff++} experience sharp performance drops outside of FF++. Xception, for example, drops from $85.32\%$ ACC on FF++ to below $60\%$ on most other datasets, showing poor generalizability. SLADD performs better in R but lacks overall consistency in F1 and AUC. This is partly explained by their weak score separability, \ie, both models show low polarity, meaning they struggle to distinguish real and fake samples confidently. 
Even though SLADD maintains stable thresholds, its overlapping score distributions lead to unreliable decisions when tested on unfamiliar data, as shown in Figure~\textcolor{RubineRed}{5} of the main paper. 

\begin{figure}[t]
 \centering
 \begin{minipage}{0.49\linewidth}
   \centering
    \includegraphics[width=0.81\linewidth]{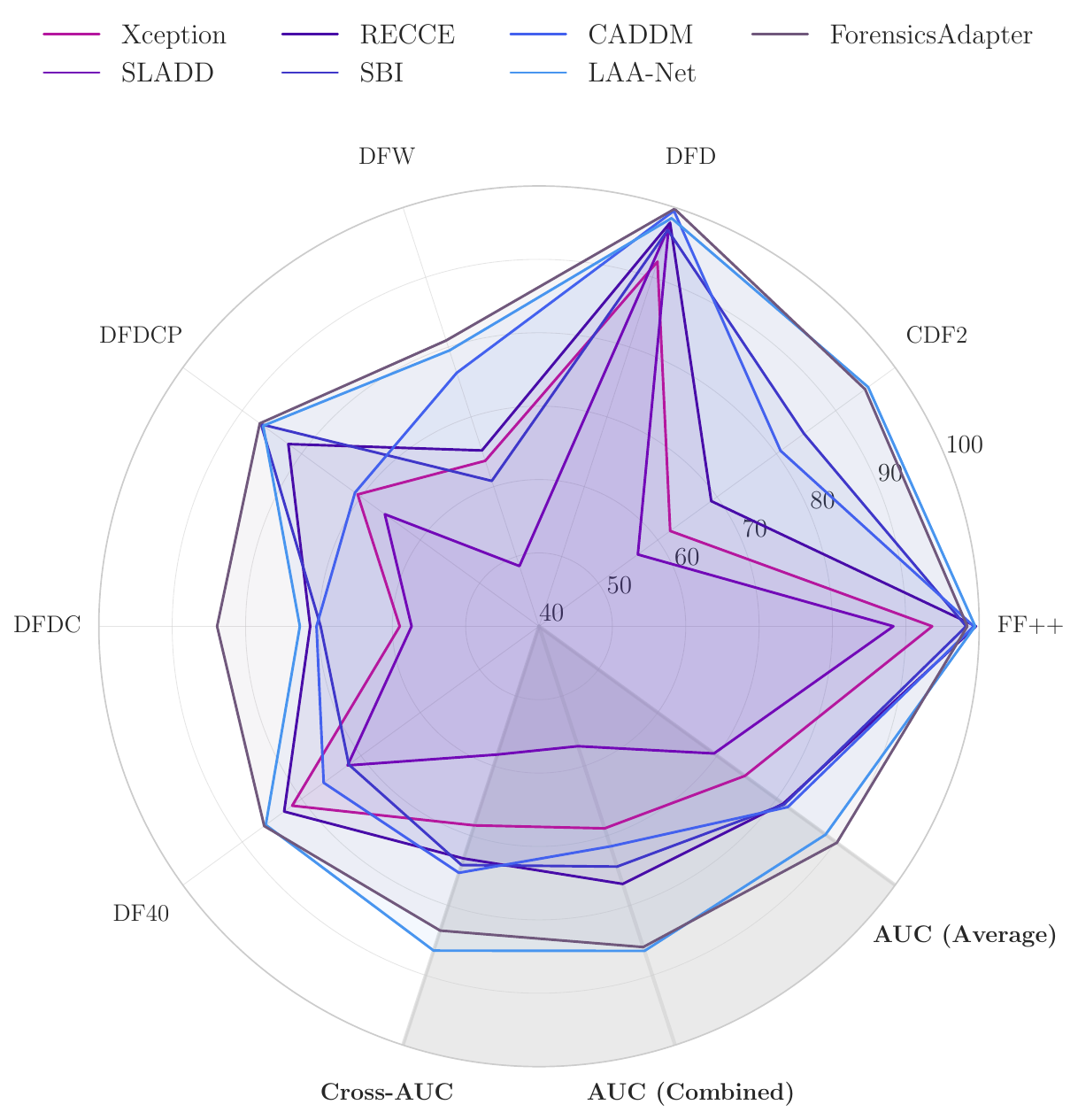}
    \vspace{-2mm}
    \caption{Methods results (AUC) for each dataset. Average AUC, combined AUC, and the proposed Cross-AUC are also reported.}
    \label{fig:radar_per_dataset}
  \end{minipage}
  \hfill
  \begin{minipage}{0.49\linewidth}
   \centering
    \includegraphics[width=0.71\linewidth]{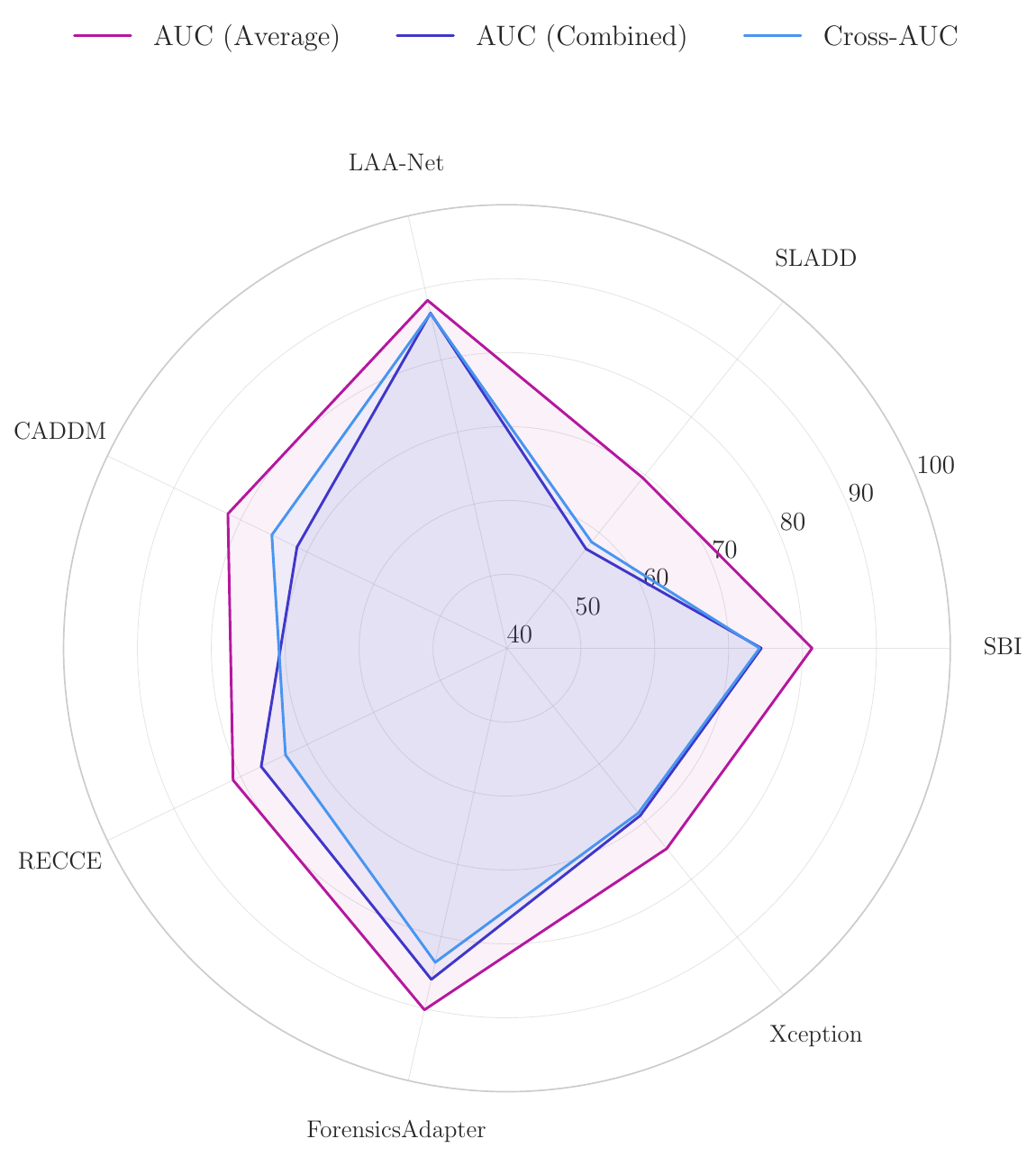}
    \vspace{-2mm}
    \caption{Direct comparison of average AUC, combined AUC, and the proposed Cross-AUC on the studied deepfake detectors.}
    \label{fig:radar_per_method}
  \end{minipage}
 \vspace{-5mm}
\end{figure}

In particular, the AUC trends highlight these generalization patterns more clearly. ForensicAdapter is the only model that maintains AUC above $80\%$ across all test datasets, indicating consistent class separation under domain shifts. Other models, including RECCE, LAA-Net, SBI, and CADDM, perform well on specific datasets but show noticeable drops on others, reflecting varying sensitivity to the target domain. Xception achieves competitive results on datasets like DFD and DF40, though its performance varies more widely across others. Notably, SLADD shows the most limited generalization, with the lowest Combined AUC and consistently low polarity scores, suggesting that it struggles to separate real and fake samples reliably across domains. A visual comparison of performance between detectors is shown in Figure~\ref{fig:radar_per_dataset}.

\vspace{2mm}
\noindent\textbf{The Combined Test Set: A Realistic Generalization Benchmark.}
As discussed in the main paper, we additionally propose a ``Combined'' test set, created by merging predictions from all test datasets, to reflect real-world use cases in which content is highly mixed. It offers a challenging test of generalization, \ie, models are exposed to many manipulation types, compression levels, and domain shifts simultaneously.

ForensicAdapter again stands out with high Combined F1 ($77.86\%$), low EER ($22.39\%$), and strong AUC ($85.96\%$). Its performance is both accurate and stable across conditions. LAA-Net reaches the highest Combined AUC ($86.51\%$) and excels in P and S but suffers from low R, leading to a lower Combined F1 ($52.25\%$). RECCE records the highest R and F1, exhibiting consistent behavior even in the most varied testing condition.
In contrast, models like SLADD, SBI, and Xception show noticeable performance drops on the Combined set. SLADD’s high R leads to many false positives, while SBI and Xception exhibit more conservative behavior or limited score separability, affecting their reliability under mixed-domain conditions.

This comparison highlights the limitation of using per-dataset averages to assess generalization. While averages may suggest overall stability, they can mask poor performance on specific domains. For example, SBI shows a decent average AUC but fails on DFDC due to very low R. SLADD has high average R, but low S on several datasets leads to frequent false positives. In contrast, ForensicAdapter performs consistently across both average and Combined evaluations, confirming its robustness. Hence, the Combined test set provides a more realistic and reliable measure of performance under diverse conditions.

\vspace{2mm}
\noindent\textbf{Cross-AUC versus Average AUC: A Visual Comparison.}
In addition to the results reported in Table~\textcolor{RubineRed}{3} of the main paper as well as discussed in Section~\textcolor{RubineRed}{4.2} of the main paper, we provide in Figure~\ref{fig:radar_per_method} a radar-based comparison between our proposed Cross-AUC and the conventional average AUC. As shown, across all models, Cross-AUC aligns more closely with \(\text{AUC}_c\) than \(\text{AUC}_a\), yielding the smallest difference (\(\Delta_\text{Cross-AUC}\)) in all cases. This underscores the importance of prediction polarization for assessing the generalization capability of detectors in mixed-domain and unconstrained settings.

\vspace{2mm}
\noindent\textbf{Threshold Stability and Score Separability.}
We examine threshold variability using $\phi_\tau$ and $\tau$, which reflect how consistently a model applies its decision boundary and how confident it is in its predictions.

It can be observed that RECCE and SLADD show the most stable threshold behavior. SLADD has identical $\phi_\tau$ values of $0.0001$ across all datasets, and RECCE maintains similarly low values between $0.00001$ and $0.0002$. Their $\tau$ values are also consistent. RECCE operates at higher $\tau$ ($0.77$ on average), indicating more cautious predictions. This stability is supported by its moderately high polarity on the Combined set ($W_{\text{Q}} = 0.232$), suggesting reasonably well-separated score distributions. SLADD, on the other hand, has the lowest polarity ($W_{\text{Q}} = 0.014$), showing that even with stable thresholds, the model struggles to confidently separate real and fake predictions.

SBI and Xception also show low $\phi_\tau$ values, but their lower polarity scores suggest that threshold consistency does not translate into well-separated scores. Their decisions are more stable than confident. ForensicAdapter shows slightly more variation in threshold placement (\eg, $0.0027$ on DFD), including the highest $\phi_\tau$ on the Combined set ($0.0007$), but its high polarity ($W_{\text{Q}} = 0.336$) supports confident decisions despite small shifts in threshold.

LAA-Net and CADDM show less consistent threshold behavior, with noticeably higher $\phi_\tau$ values on FF++ ($0.0161$ and $0.0253$), and more fluctuation in $\tau$ across datasets. While both achieve high polarity, the instability in thresholds suggests more erratic decision behavior under domain shifts.

Overall, RECCE is the only model that combines both stable thresholds and clear score separability. Other models show strength in one aspect but not both, which may impact their reliability under unseen conditions.

\vspace{2mm}
\noindent\textbf{Score Polarization and Its Relationship to Classification Metrics.}
\label{subsec:polarization_metrics}
We analyze polarity $W$ using WD as $\Pi$ between the score distributions of real and fake samples to quantify how well a model separates the two classes. Since classification metrics like AUC, EER, and F1 are directly influenced by how confidently a model scores real versus fake data, we expect models with higher polarity to generally exhibit stronger performance. In our evaluation, we examine how this relationship holds across models under cross-domain conditions.

On the Combined test set, we observe that models with higher polarity, such as ForensicAdapter ($W_{\text{Q}} = 0.336$), LAA-Net ($W_{\text{Q}} = 0.318$), and RECCE ($W_{\text{Q}} = 0.232$), also achieve strong classification results. ForensicAdapter pairs the highest polarity with an AUC of $85.96\%$ and F1 of $83.49\%$, indicating that well-separated score distributions contribute directly to confident and reliable decisions. RECCE follows a similar trend, where moderate-to-high polarity aligns with high R and competitive F1, further supported by its stable thresholds across domains. LAA-Net shows high polarity and the highest AUC ($86.51\%$), which suggests good score separation. However, its R remains low, leading to a reduced F1 score ($52.25\%$). This gap indicates that while the model distinguishes real and fake scores well, it tends to bias predictions toward the real class, possibly due to misaligned threshold behaviors or domain sensitivity. 

In contrast, SLADD and Xception exhibit much lower polarity, with $W_{\text{Q}} = 0.014$ and $0.190$ respectively. SLADD shows stable thresholds across datasets, but the score overlap between real and fake classes results in high false positive rates and low F1 on the Combined set. This confirms that stable thresholds alone do not guarantee good performance. Without clear separation in the score space, the models' predictions become sensitive and less confident under domain shifts.
Xception, while having slightly higher polarity than SLADD, struggles with both generalization and score separability. Its performance drops sharply outside of FF++, and the weak polarization across domains contributes to high EER and poor generalizability.
SBI, meanwhile, offers a different failure mode. Despite modest polarity ($W_{\text{Q}} = 0.172$), it achieves very high P and S, but extremely low R. Its behavior reflects the use of conservative thresholds rather than confident score separation, a pattern supported by its relatively low $\tau$ values and narrow decision margins. While this approach minimizes false positives, it also prevents the model from detecting many fake samples, limiting its utility in general-purpose settings.

These findings confirm that polarity is a strong indicator of how confidently a model distinguishes real and fake content, and it typically correlates with AUC and EER. However, it must be considered alongside threshold behavior and precision-recall balance to fully understand a model’s generalization capability. Models that combine high polarity with consistent thresholds and balanced decision strategies, such as ForensicAdapter and RECCE, are more reliable under cross-domain conditions.

\vspace{2mm}
\noindent\textbf{Precision-Recall Trade-off.}
While earlier sections analyze model performance from score separability and threshold stability perspectives. In this section, we focus on the practical trade-offs between P and R. Models like RECCE and SLADD tend to favor R, detecting more fake samples but at the cost of increased false positives. In contrast, SBI and LAA-Net show conservative behavior, prioritizing P and S while often missing actual deepfakes. ForensicAdapter offers the most balanced trade-off, maintaining high values for both P and R. These differences highlight how models vary not just in accuracy, but also in how cautiously or aggressively they make decisions, an important consideration for real-world deployment.

\vspace{2mm}
\noindent\textbf{Failure Case Analysis.}
We observe consistent performance drops on certain datasets, revealing where generalization tends to fail. These cases highlight the sensitivity of some models to domain-specific shifts not reflected in overall averages.
As shown in Table~\ref{tabl:full_res}, the DFDC dataset poses a common challenge. Models such as LAA-Net, RECCE, and CADDM show significant declines in performance. For example, LAA-Net drops to $16.33\%$ R and $27.69\%$ F1, despite achieving top AUCs on other datasets. RECCE and CADDM also report high EERs ($34.43\%$ and $36.06\%$) and reduced F1, indicating difficulty adapting to DFDC’s unique characteristics, such as higher compression or different manipulation styles.
DF40 also reveals failure points. Xception and SBI, in particular, suffer from very low R ($14.44\%$ and $21.46\%$, respectively), despite maintaining high S. These patterns suggest that overly conservative models fail to detect fakes when the score distributions shift, even if their decision thresholds remain stable.

Such failures emphasize the limits of models that rely heavily on training distribution priors. While models like ForensicAdapter maintain robust results even under domain variation, others show clear breakdowns when faced with unseen manipulation types or domain properties not covered by FF++.

\begin{table}[]
\centering
\vspace{-5mm}
\caption{Runtime (in seconds) for computing polarization using different estimators across datasets.}
\vspace{-3mm}
\resizebox{0.72\linewidth}{!}{
\begin{tabular}{c|cccccccc}
    \toprule
         & FF++ & CDF & DFD & DFW & DFDCP & DFDC & DF40 & Combined \\
         \midrule
         \midrule
         $W_{\text{KDE}}$ & 0.0014 & 0.0012 & 0.0012 & \boldgreen{0.0010} & 0.0012 & 0.0015 & 0.0023 & 0.0036 \\
         $W_{\text{BD}}$ & 0.0013 & 0.0012 & 0.0011 & 0.0011 & 0.0014 & \boldgreen{0.0011} & \boldgreen{0.0012} & \boldgreen{0.0013} \\
         $W_{\text{Q}}$ & \boldgreen{0.0008} & \boldgreen{0.0008} & \boldgreen{0.0008} & 0.0013 & \boldgreen{0.0007} & 0.0129 & 0.0148 & 0.0847 \\
         $W_{\text{GMM}}$ & 2.2309 & 2.1118 & 2.2576 & 2.7162 & 2.0931 & 4.8132 & 5.2694 & 8.3977 \\
         \bottomrule
\end{tabular}
}
\label{tab:runtime}
\vspace{-6mm}
\end{table}

\vspace{2mm}
\noindent\textbf{Runtime Analysis.}
Table~\ref{tab:runtime} reports the runtime required to compute polarization for each estimator. As shown, non-parametric variants ($W_{\text{KDE}}$, $W_\text{Q}$) incur only millisecond-level overhead, while the GMM-based estimator ($W_{\text{GMM}}$) is slower. Nevertheless, all variants remain practical for offline evaluation. Given its strong empirical performance and low computational cost, we use the quantile-based WD estimator ($W_Q$) by default.

\begin{table}[]
    \centering
    \vspace{-4mm}
    \caption{Effect of the number of test domains.}
    \vspace{-3mm}
    \resizebox{0.66\linewidth}{!}{
    \begin{tabular}{c|c|c|cc}
        \toprule
         No. of Domains & Method & $\text{AUC}_c$ & $\text{AUC}_a$ & $\text{Cross-AUC}$ \\
         \midrule
            & Xception~\cite{ff++} & 68.95	& 74.68 & \boldgreen{68.52} \\
            & SLADD~\cite{sladd} & 57.18 & 69.48 & \boldgreen{58.37} \\
            & RECCE~\cite{ete_recons} & 76.91 & 81.13 & \boldgreen{73.26} \\
    7 (All) & SBI~\cite{sbi} & 74.43 & 81.30 & \boldgreen{74.20} \\
            & CADDM~\cite{caddm} & 71.53 & 81.91 & \boldgreen{75.31} \\
            & LAA-Net~\cite{laa_net} & 86.51 & 88.28 & \boldgreen{86.47} \\
            & ForensicsAdapter~\cite{forensicsadapter} & 85.96 & 90.18 & \boldgreen{83.59} \\
        \midrule
            & Xception~\cite{ff++} & 65.13 & 73.53 & \boldgreen{68.53} \\
            & SLADD~\cite{sladd} & 62.55 & 69.02 & \boldgreen{57.62} \\
            & RECCE~\cite{ete_recons} & 74.16 & 80.83 & \boldgreen{72.70} \\
6 (w/o DF40)& SBI~\cite{sbi} & 75.62 & 82.84 & \boldgreen{75.38} \\
            & CADDM~\cite{caddm} & 76.42 & 82.85 & \boldgreen{76.88} \\
            & LAA-Net~\cite{laa_net} & 84.79 & 88.66 & \boldgreen{86.33} \\
            & ForensicsAdapter~\cite{forensicsadapter} & 87.36 & 90.83 & \boldgreen{85.20} \\
        \midrule
            & Xception~\cite{ff++} & 73.32 & 76.44 & \boldgreen{73.06} \\
            & SLADD~\cite{sladd} & 67.58 & 71.35 & \boldgreen{64.09} \\
            & RECCE~\cite{ete_recons} & 76.76 & 82.76 & \boldgreen{78.51} \\
5 (w/o DF40, DFDC) & SBI~\cite{sbi} & 82.19 & 85.46 & \boldgreen{80.81} \\
            & CADDM~\cite{caddm} & 83.13 & 85.35 & \boldgreen{82.60} \\
            & LAA-Net~\cite{laa_net} & 94.52 & 91.87 & \boldgreen{93.77} \\
            & ForensicsAdapter~\cite{forensicsadapter} & 91.45 & \boldgreen{92.22} & 93.53 \\
        \midrule
            & Xception~\cite{ff++} & 74.58 & 77.91 & \boldgreen{73.56} \\
            & SLADD~\cite{sladd} & 68.31 & 72.70 & \boldgreen{65.87} \\
            & RECCE~\cite{ete_recons} & 81.91 & \boldgreen{82.89} & 79.33 \\
4 (w/o DF40, DFDC, DFDCP) & SBI~\cite{sbi} & 82.09 & 85.12 & \boldgreen{81.36} \\
            & CADDM~\cite{caddm} & 85.76 & 88.94 & \boldgreen{87.57} \\
            & LAA-Net~\cite{laa_net} & 95.72 & 93.22 & \boldgreen{93.25} \\
            & ForensicsAdapter~\cite{forensicsadapter} & 92.75 & 93.52 & \boldgreen{92.12} \\
        \bottomrule
    \end{tabular}
    }
    \label{tab:no_domains}
\vspace{-4mm}
\end{table}

\vspace{2mm}
\noindent\textbf{Effect of the Number of Test Domains.}
In Table~\ref{tab:no_domains}, we conduct an ablation study by progressively reducing the number of test domains from seven to four and measuring how Cross-AUC, average AUC ($\text{AUC}_a$), and Combined AUC ($\text{AUC}_c$) varies across different detectors. We found that Cross-AUC consistently remains close to $\text{AUC}_c$, even as domain diversity decreases, whereas $\text{AUC}_a$ tends to overestimate the performance. This supports the stability of Cross-AUC under varying domain coverage, suggesting that it reflects real-world generalization risk more reliably than per-domain average.


\begin{table}[]
    \centering
    \vspace{-4mm}
    \caption{Detailed polarization scores for different estimators, which are used to compute the Cross-AUC results reported in Table~\textcolor{RubineRed}{6} of the main paper.}
    \vspace{-2mm}
    \resizebox{0.9\linewidth}{!}{
    \begin{tabular}{c|c|ccccccccc}
        \toprule
        \multirow{2}{*}{Method} & \multirow{2}{*}{Estimator} & \multicolumn{9}{c}{Polarization Score} \\ \cline{3-11}
        & & FF++ & CDF & DFD & DFW & DFDCP & DFDC & DF40 & Average & Combined \\
        \midrule
        \midrule
        \multirow{3}{*}{Xception~\cite{ff++}} & WD & 0.661 & 0.077 & 0.524 & 0.139 & 0.183 & 0.091 & 0.264 & 0.277 & 0.190 \\
            & KL & 23.597 & 17.919 & 16.039 & 15.340 & 20.217 & 4.925 & 11.431 & 15.638 & 2.999 \\
            & JS & 0.797 & 0.712 & 0.794 & 0.699 & 0.792 & 0.413 & 0.554 & 0.680 & 0.345 \\
        \midrule
        \multirow{3}{*}{SLADD~\cite{sladd}} & WD & 0.079 & 0.005 & 0.090 & 0.002 & 0.015 & 0.010 & 0.026 & 0.032 & 0.014 \\
            & KL & 17.695 & 4.914 & 21.158 & 2.073 & 14.469 & 0.553 & 2.764 & 9.089 & 0.744 \\
            & JS & 0.686 & 0.374 & 0.767 & 0.303 & 0.593 & 0.199 & 0.375 & 0.471 & 0.185 \\
        \midrule
        \multirow{3}{*}{RECCE~\cite{ete_recons}} & WD & 0.716 & 0.136 & 0.355 & 0.115 & 0.243 & 0.203 & 0.144 & 0.273 & 0.232 \\
            & KL & \boldgreen{26.704} & 17.120 & 4.266 & 12.087 & 20.226 & 4.199 & 1.922 & 12.361 & 1.187 \\
            & JS & 0.824 & 0.701 & 0.755 & 0.628 & 0.791 & 0.464 & 0.518 & 0.669 & 0.418 \\
        \midrule
        \multirow{3}{*}{SBI~\cite{sbi}} & WD & 0.576 & 0.238 & 0.577 & 0.085 & 0.244 & 0.108 & 0.116 & 0.278 & 0.172 \\
            & KL & 21.895 & 18.545 & 22.352 & 9.984 & 20.633 & 5.346 & 6.802 & 15.080 & 5.859 \\
            & JS & 0.813 & 0.738 & 0.813 & 0.557 & 0.809 & 0.381 & 0.426 & 0.648 & 0.389 \\
        \midrule
        \multirow{3}{*}{CADDM~\cite{caddm}} & WD & 0.726 & 0.226 & 0.466 & 0.246 & 0.179 & 0.174 & 0.151 & 0.310 & 0.198 \\
            & KL & 25.675 & 18.646 & \boldgreen{24.948} & 15.572 & 19.638 &	5.993 & 7.800 & 16.896 & 3.873 \\
            & JS & \boldgreen{0.830} & 0.736 & 0.825 & 0.673 & 0.773 & 0.441 & 0.492 & 0.681 & 0.365 \\
        \midrule
        \multirow{3}{*}{LAA-Net~\cite{laa_net}} & WD & \boldgreen{0.927} & \boldgreen{0.582} & 0.708 & \boldgreen{0.285} & \boldgreen{0.446} & 0.156 & 0.299 & \boldgreen{0.486} & 0.318 \\
            & KL & 25.423 & 21.371 & 23.247 & \boldgreen{19.313} & 19.500 & 6.680 & 12.617 & 18.307 & 9.745 \\
            & JS & 0.822 & \boldgreen{0.800} & 0.804 & \boldgreen{0.742} & 0.749 & 0.397 & 0.562 & 0.696 & 0.528 \\
        \midrule
        \multirow{3}{*}{ForensicsAdapter~\cite{forensicsadapter}} & WD & 0.540 & 0.374 & \boldgreen{0.713} & \boldgreen{0.285} & 0.293 & \boldgreen{0.286} & \boldgreen{0.322} & 0.402 & \boldgreen{0.336} \\
            & KL & 23.222 & \boldgreen{21.466} & 24.085 & 17.018 & \boldgreen{20.781} & \boldgreen{11.225} & \boldgreen{13.890} & \boldgreen{18.812} & \boldgreen{10.989} \\
            & JS & 0.819 & 0.798 & \boldgreen{0.830} & 0.708 & \boldgreen{0.815} & \boldgreen{0.557} & \boldgreen{0.590} & \boldgreen{0.731} & \boldgreen{0.535} \\
        \bottomrule
    \end{tabular}
    }
    \label{tab:full_polarization}
    \vspace{-4mm}
\end{table}

\vspace{2mm}
\noindent\textbf{Extended Analysis: Cross-AUC with Different Polarization Estimators.}
As discussed in Section~\textcolor{RubineRed}{4.2} of the main paper, to assess the influence of the polarization component, we evaluate Cross-AUC with three different estimators including Wasserstein Distance (WD), KL Divergence (KL), and Jensen-Shannon Distance (JS). Table~\textcolor{RubineRed}{6} of the main paper reports the resulting Cross-AUC values, while Table~\ref{tab:full_polarization} in this supplementary material details the underlying polarization scores from each estimator. The results show that both WD and KL yield polarization magnitudes that scale proportionally with the degree of domain mismatch, leading to Cross-AUC values that remain close to the combined AUC. In contrast, JS produces less discriminative polarization scores, which translate into weaker alignment in Cross-AUC. This suggests that although Cross-AUC is generally robust to the choice of estimator, the WD formulation offers the most stable and interpretable behavior, with KL as a competitive alternative.


\begin{table}
\label{tabl:orig_data_dis}
\caption{Overview of datasets. 
The number of ``Methods'' is categorized into four subsets: Face-Swapping (FS), Face-Reenactment (FR), Entire Face Synthesis (EFS), and Face Editing (FE). ``Perturbs'' denotes the number of perturbations applied, while ``Unknown'' indicates whether the source contains prior knowledge of manipulations.}
\centering
\vspace{-2mm}
\resizebox{0.9\linewidth}{!}{
\begin{tabular}{c|c|cc|cccc|c|c|c}
\toprule
Dataset & Venue & Real videos & Fake videos & FS & FR & EFS & FE & Methods & Perturbs & Unknown \\
\midrule
\midrule
FF++~\cite{ff++} & ICCV'19 & 1,000 & 4,000 & 2 & 2 & - & - & 4 & 2 & $\times$ \\
DFD~\cite{dfd} & None & 363 & 3,068 & 5 & - & - & - & 5 & - & $\times$ \\
DFDCP~\cite{dfdcp} & ArXiv'19 & 1,131 & 4,113 & - & - & - & - & 2 & 3 & $\times$ \\
CDF~\cite{celeb_df} & CVPR'20 & 590 & 5,639 & 1 & - & - & - & 1 & - & $\times$ \\
DFW~\cite{wdf} & ACMMM'20 & 3,805 & 3,509 & - & - & - & - & - & - & \checkmark \\
DFDC~\cite{dfdc} & ArXiv'20 & 23,564 & 104,500 & 5 & 1 & 2 & - & 8 & 19 & $\times$ \\
DF40~\cite{DF40} & NeurIPS'24 & 1428 & 0.1M+ & 10 & 13 & 12 & 5 & 40 & 6 & $\times$ \\
\bottomrule
\end{tabular}
}
\label{tabl:overview_datasets}
\vspace{-4mm}
\end{table}

\subsection{More Details of Datasets}
\label{subsec:dataset_info}

Seven standard and challenging datasets are used in our experiments, including \textbf{FaceForensics}++ (FF++)~\cite{ff++}, \textbf{Celeb-DF} (CDF)~\cite{celeb_df}, \textbf{Google Deepfake Detection} (DFD)~\cite{dfd}, \textbf{WildDeepfake} (DFW)~\cite{wdf}, \textbf{Deepfake Detection Challenge} (DFDC)~\cite{dfdc}, and \textbf{Deepfake Detection Challenge Preview} (DFDCP)~\cite{dfdcp}, and \textbf{DF40}~\cite{DF40}. 
These datasets are selected based on the following criteria: 1) they are widely used for evaluation in the research community; 2) they contain fakes from diverse sources, including those generated by black-box or undisclosed methods, for which no implementation details or prior knowledge are available; 3) they present increased complexity as a result of various image perturbations, such as compression, noise, or blur. An overview of the selected benchmarks is presented in Table~\ref{tabl:overview_datasets}.

Specifically, FF++ is typically used for training, while the others serve as test sets in the cross-dataset evaluation protocol. It consists of $1000$ real videos and $4000$ fakes videos with different compression levels, which are generated by four manipulation techniques, i.e., Deepfakes (DF)~\cite{deepfake}, Face2Face (F2F)~\cite{face2face}, FaceSwap (FS)~\cite{faceswap}, and NeuralTextures (NT)~\cite{neutex}. In our experiments, we adopt the raw version of FF++. CDF is one of the most common and challenging benchmarks typically used to assess the generalizability of deepfake detectors under the cross-dataset evaluation setting~\cite{fxray, ict, ete_recons, sladd, sbi, aunet, laa_net, forensicsadapter}. It contains highly realistic deepfakes. DFD includes more than $3000$ forged videos featuring $28$ actors in various scenes. DFW is fully sourced from the internet, without prior knowledge of generation methods. DFDC and its preview version (DFDCP) are challenging, large-scale datasets that contain numerous distorted videos applied by several perturbation techniques such as compression, noise, etc. The recent DF40 benchmark, a highly diverse and large-scale dataset comprising 40 distinct deepfake techniques, enables more comprehensive evaluations for the next generation of deepfake detection.

\subsection{More Details of Deepfake Detectors}
\label{subsec:detectors}

In this section, we describe the seven state-of-the-art (SOTA) deepfake detection models~\cite{xception, sladd, ete_recons, sbi, caddm, laa_net, forensicsadapter} evaluated under our proposed protocol (see Section~\textcolor{RubineRed}{4.1} in the main paper).

\vspace{1mm}
\noindent\textbf{XceptionNet~\footnote{\url{https://github.com/ondyari/FaceForensics}}: }
In this work, the well-known benchmark FF++~\cite{ff++} is introduced. The Xception architecture proposed in~\cite{xception} was trained in an end-to-end manner using both real and fake data from FF++~\cite{ff++}.


\vspace{1mm}
\noindent\textbf{SLADD~\footnote{\url{https://github.com/liangchen527/SLADD}}: }
In~\cite{sladd}, the authors introduce an approach called SLADD~\cite{sladd} that leverages adversarial training~\cite{adver_training, adver_training_scale, adver_augment}.  In particular, it relies on two components: a generator and a discriminator (detector). While the generator attempts to synthesize facial images based on splicing operations, the detector aims to discriminate between fake and real faces. SLADD is trained using the generated pseudo-fakes as well as real and deepfake images from FF++. This approach can be seen as a hybrid method since it simultaneously takes advantage of a data synthesis strategy as well as a multi-task learning framework. This approach is claimed to be generic, given the high AUC achieved under the cross-dataset protocol.

\vspace{1mm}
\noindent\textbf{RECCE~\footnote{\url{https://github.com/VISION-SJTU/RECCE}}: }
This method, referred to as RECCE~\cite{ete_recons}, is based on a reconstruction-classification schema. While the reconstruction learning over real images allows for extracting forgery-aware features, the classification learning mines the main discrepancy between real and fake images. Both real and fake data from FF++~\cite{ff++} are used during training. This multi-task learning framework is shown to achieve competitive AUC performance under the cross-dataset setting.

\vspace{1mm}
\noindent\textbf{SBI~\footnote{\url{https://github.com/mapooon/SelfBlendedImages}}: }
This approach, called SBI~\cite{sbi}, is based on a novel data synthesis. This data synthesis produces pseudo-fakes by blending source and target images from a single pristine image. The main idea is that such an augmentation generates pseudo-fake images with very subtle artifacts, thereby pushing the classifier to learn more generic representations. Only real data from  FF++~\cite{ff++} and pseudo-fakes are used for training a binary classifier. Similar to~\cite{ete_recons}, an impressive AUC is registered for the cross-dataset setting.

\vspace{1mm}
\noindent\textbf{CADDM~\footnote{\url{https://github.com/megvii-research/CADDM}}:} The proposed method, ID-unaware Deepfake Detection Model, aims to improve the generalization of deepfake detectors by addressing the issue of Implicit Identity Leakage (IIL), the unintended reliance on identity-specific features. To mitigate this, the authors introduce an Artifact Detection Module (ADM) that encourages the model to focus on local visual artifacts rather than global identity features. ADM uses a multi-scale anchor-based detection approach to identify regions in an image that may contain manipulation traces, helping the model distinguish between real and fake content at a finer, more localized level. To support ADM training, they design a Multi-scale Facial Swap (MFS) technique that generates fake images with labeled artifact areas by blending regions of source and fake images using different window sizes and blending strategies. This synthetic data provides ground-truth artifact locations, allowing the model to learn more generalized features. By shifting attention away from identity and toward manipulation cues, this method significantly improves cross-dataset performance and robustness to unseen forgery techniques.

\vspace{1mm}
\noindent\textbf{LAA-Net~\footnote{\url{https://github.com/10Ring/LAA-Net}}:} LAA-Net (Localized Artifact Attention Network) is a fine-grained deepfake detection framework designed to effectively detect high-quality and unseen manipulations by focusing on localized visual inconsistencies. At its core, LAA-Net introduces an explicit attention mechanism within a multi-task learning framework composed of three parallel branches: a binary classification branch to distinguish real from fake images, a heatmap branch to localize regions with potential blending artifacts guided by ``vulnerable points'', and a self-consistency branch that evaluates similarity between vulnerable pixels and their surroundings. These branches are trained using pseudo-fake data generated through blending-based synthesis from real images, eliminating the need for large-scale fake datasets. Additionally, LAA-Net integrates an Enhanced Feature Pyramid Network (E-FPN) to capture and preserve discriminative low-level features across multiple scales while minimizing redundancy. By combining precise attention guidance with robust feature representation, LAA-Net significantly improves the generalizability and accuracy of deepfake detection, even under challenging cross-dataset and quality-agnostic scenarios.

\vspace{1mm}
\noindent\textbf{ForensicAdapter~\footnote{\url{https://github.com/OUC-VAS/ForensicsAdapter}}:} ForensicsAdapter is a lightweight and task-specific adapter network designed to transform CLIP into a powerful and generalizable face forgery detector. Unlike previous CLIP-based approaches that treat CLIP as a frozen feature extractor, this method introduces a dedicated adapter module that operates in parallel with CLIP to detect subtle manipulation traces, especially blending boundaries characteristic of forged faces. The adapter is trained using a combination of objectives: masked blending boundary detection, patch-wise contrastive learning, and sample-wise contrastive learning, each tailored to enhance sensitivity to forgery cues. A novel interaction mechanism is also proposed to enable bidirectional knowledge exchange between the adapter and CLIP: the adapter absorbs low-level CLIP visual tokens to enrich its learning, and in return, uses attention bias to guide CLIP's focus toward forgery-specific features without altering CLIP’s core parameters. Despite having only $5.7M$ trainable parameters, ForensicsAdapter achieves state-of-the-art performance across multiple benchmarks and proves highly robust to domain shifts and image perturbations, making it a strong baseline for future CLIP-based forgery detection work.

\subsection{Additional Evaluation Metrics}
\label{subsec:metrics}

In addition to score polarization measured by Wasserstein Distance ($W$), AUC, and our proposed Cross-AUC, we report additional metrics to support a more in-depth analysis. These include Accuracy (ACC), Balanced Accuracy (BACC), Precision (P), Recall (R), Specificity (S), F1-score (F1), and Equal Error Rate (EER). We consider the \textit{Fake} and \textit{Real} classes as the \textit{Positive} and \textit{Negative} classes, respectively.

\vspace{1mm}
\noindent\textbf{Equal Error Rate (EER):} is a common metric for evaluating the performance of biometric authentication systems, such as speaker verification or facial recognition. It represents the point at which the False Acceptance Rate (FAR) and the False Rejection Rate (FRR) are equal. At this threshold, the system is equally likely to incorrectly accept an unauthorized user (false positive) as it is to incorrectly reject an authorized user (false negative). A lower EER indicates better overall system performance, as it reflects a more optimal trade-off between FAR and FRR.

In the context of Deepfake Detection, EER is used to evaluate the ability of a detector to distinguish between real and manipulated (fake) content. Here, the FAR corresponds to the rate at which fake content is incorrectly classified as real, and the FRR corresponds to the rate at which real content is incorrectly classified as fake. EER provides a single value that reflects the detector's robustness and balance between these two types of errors.

\vspace{1mm}
\noindent\textbf{Accuracy (ACC): }
It is computed based on the percentage of correct predictions as follows,
\begin{equation}
   \text{ACC} = \frac{\text{TP}+\text{TN}}{\text{TP}+\text{FP}+\text{TN}+\text{FN}} \text{,}
\end{equation}
where TP, FP, TN, and FN denote True Positives, False Positives, True Negatives, and False Negatives, respectively.

While commonly used, ACC can be misleading when the dataset is imbalanced~\cite{AUC_maxi, partial_auc_maxi}. For instance, in the context of deepfake detection, standard datasets such as FF++ contain four times more fake than real samples (see Table~\ref{tabl:overview_datasets}). As a result, a model biased toward predicting the majority class (Fake) may achieve high ACC without truly distinguishing between real and fake content. To address this, we also report Balanced Accuracy (BACC), which provides a more informative view under class imbalance.

\vspace{1mm}
\noindent\textbf{Balanced Accuracy (BACC): }
It is calculated based on the average recall values for both positive and negative classes.
\begin{equation}
    \text{BACC} = \frac{1}{2} \left( \frac{\text{TP}}{\text{TP} + \text{FN}} + \frac{\text{TN}}{\text{TN} + \text{FP}} \right) \text{.}
\end{equation}

\vspace{1mm}
\noindent\textbf{Specificity (S):} It describes how well the model detects the negative samples. It is calculated as the proportion of correct negative predictions over the actual total number of negative samples. The formula is defined below,

\begin{equation}
      \text{S} = \frac{ \text{TN}}{ \text{TN}+ \text{FP}}\text{.}
\end{equation}

\vspace{1mm}
\noindent\textbf{Precision (P): }
It describes how correct the positive predictions are. It is calculated as the proportion of correctly predicted positive samples over the total number of predictions classified as positive. The formula is defined below,

\begin{equation}
    \text{P} = \frac{ \text{TP}}{ \text{TP}+ \text{FP}} \text{.}
\end{equation}

\vspace{1mm}
\noindent\textbf{Recall (R): }
It describes how well the model detects the positive samples. It is calculated as the proportion of correct positive predictions over the actual total number of positive samples. The formula is defined below,

\begin{equation}
      \text{R} = \frac{ \text{TP}}{ \text{TP}+ \text{FN}}\text{.}
\end{equation}

\vspace{1mm}
\noindent\textbf{F1-score (F1): } It is often used to assess the overall performance of a classification model. It is computed as follows,

\begin{equation}
    \text{F1} = 2.\frac{\text{P}.\text{R}}{\text{P}+\text{R}} \text{.}
\end{equation}

\end{document}